Florentin Smarandache

# INTRODUCTION TO NEUTROSOPHIC MEASURE, NEUTROSOPHIC INTEGRAL, AND NEUTROSOPHIC PROBABILITY

*Many books can be downloaded from the following*

***Digital Library of Science***:
*http://fs.gallup.unm.edu/eBooks-otherformats.htm*

## C O N T E N T S

**Chapter 3.**

# Neutrosophic Science

*(Preface)*

*Since the world is full of indeterminacy, the neutrosophics found their place into contemporary research.*

*We now introduce for the first time the notions of* **neutrosophic measure** *and* **neutrosophic integral***.*
*We develop the 1995 notion of* **neutrosophic probability** *and give many practical examples.*

*Neutrosophic Science means development and applications of neutrosophic logic/set/measure/integral/probability etc. and their applications in any field.*

*It is possible to define the neutrosophic measure and consequently the neutrosophic integral and neutrosophic probability in many ways, because there are various types of indeterminacies, depending on the problem we need to solve.*
*Indeterminacy is different from randomness. Indeterminacy can be caused by physical space materials and type of construction, by items involved in the space, or by other factors.*

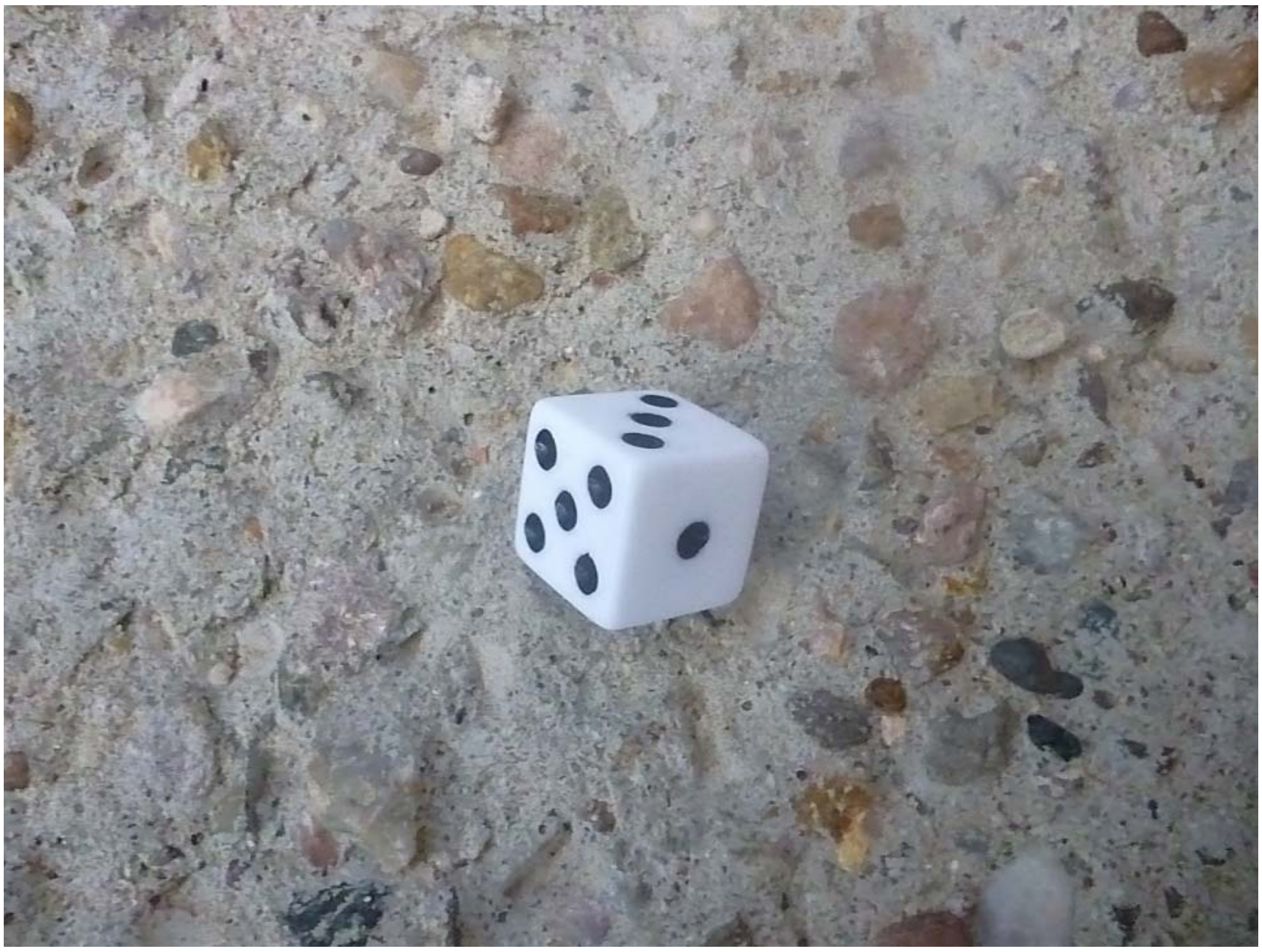

*Fig. 1. An example of indeterminacy.*
*What is tossed, 1, 3 or 5?*

*Neutrosophic measure is a generalization of the classical measure for the case when the space contains some indeterminacy.*

*Neutrosophic probability is a generalization of the classical and imprecise probabilities.*

*Several classical probability rules are adjusted in the form of neutrosophic probability rules.*

*Finally, the neutrosophic probability is extended to n-valued refined neutrosophic probability.*

*The author*

# Chapter 1.
# Introduction to Neutrosophic Measure

## 1.1. Introduction.

Let **<A>** be an item. <A> can be a notion, an attribute, an idea, a proposition, a theorem, a theory, etc. And let **<antiA>** be the opposite of <A>; while **<neutA>** be neither <A> nor <antiA> but the neutral (or indeterminacy, unknown) related to <A>.

For example, if <A> = victory, then <antiA> = defeat, while <neutA> = tie game.

If <A> is the degree of truth value of a proposition, then <antiA> is the degree of falsehood of the proposition, while <neutA> is the degree of indeterminacy (i.e. neither true nor false) of the proposition.

Also, if <A> = voting for a candidate, <antiA> = voting against that candidate, while <neutA> = not voting at all, or casting a blank vote, or casting a black vote.

In the case when <antiA> does not exist, we consider its measure be null { *m(antiA)=0* }. And similarly when <neutA> does not exist, its measure is null { *m(neutA) = 0* }.

## 1.2. Definition of Neutrosophic Measure.

We introduce for the first time the scientific notion of neutrosophic measure.

Let $X$ be a neutrosophic space, and $\Sigma$ a $\sigma$-neutrosophic algebra over $X$. A *neutrosophic measure* $\nu$ is defined by for neutrosophic set $A \in \Sigma$ by

$$\nu : X \to R^3,$$

$$\nu(A) = (m(A), m(neutA), m(antiA)), \qquad (1)$$

with *antiA* = the opposite of A, and *neutA* = the neutral (indeterminacy) neither A nor anti A (as defined above);
for any $A \subseteq X$ and $A \in \Sigma$,
*m(A)* means *measure of the determinate part of A;*
*m(neutA) means measure of indeterminate part of A;*
and *m(antiA)* means *measure of the determinate part of antiA*;
where $\nu$ is a function that satisfies the following two properties:

a) Null empty set: $\nu(\Phi) = (0, 0, 0)$.
b) Countable additivity (or $\sigma$-additivity): For all countable collections $\{A_n\}_{n \in L}$ of disjoint neutrosophic sets in $\Sigma$, one has:

$$\nu\left(\bigcup_{n \in L} A_n\right) = \left(\sum_{n \in L} m(A_n),\ \sum_{n \in L} m(neutA_n), \sum_{n \in L} m(antiA_n) - (n-1)m(X)\right)$$

where $X$ is the whole neutrosophic space, and

$$\sum_{n \in L} m(antiA_n) - (n-1)m(X) = m(X) - \sum_{n \in L} m(A_n) = m(\bigcap_{n \in L} antiA_n). \tag{2}$$

### 1.3. Neutrosophic Measure Space.

A neutrosophic measure space is a triplet $(X, \Sigma, \nu)$.

### 1.4. Normalized Neutrosophic Measure.

A neutrosophic measure is called normalized if

$$\nu(X) = (m(X), m(neutX), m(antiX)) = (x_1, x_2, x_3),$$

with $x_1 + x_2 + x_3 = 1$,

and $x_1 \geq 0, x_2 \geq 0, x_3 \geq 0$. (3)

Where, of course, $X$ is the whole neutrosophic measure space.

## 1.5. Finite Neutrosophic Measure Space.

Let $A \subset X$. We say that $\nu(A) = (a_1, a_2, a_3)$ is finite if all $a_1$, $a_2$, and $a_3$ are finite real numbers.

A *neutrosophic measure space* $(X, \Sigma, \nu)$ is called *finite* if $\nu(X) = (a, b, c)$ such that all *a, b,* and *c* are finite (rather than infinite).

## 1.6. $\sigma$-Finite Neutrosophic Measure.

A neutrosophic measure is called $\sigma$-finite if $X$ can be decomposed into a countable union of neutrosophically measurable sets of fine neutrosophic measure.

Analogously, a set $A$ in $X$ is said to have a $\sigma$*-finite neutrosophic measure* if it is a countable union of sets with finite neutrosophic measure.

## 1.7. Neutrosophic Axiom of Non-Negativity.

We say that the neutrosophic measure $\nu$ *satisfies the axiom of non-negativity*, if:

$\forall A \in \Sigma$, $\nu(A) = (a_1, a_2, a_3) \geq 0$ if $a_1 \geq 0, a_2 \geq 0$, and $a_3 \geq 0$. *(4)*

While a neutrosophic measure $\nu$, that satisfies only the null empty set and countable additivity axioms (hence not the non-negativity axiom), takes on at most one of the $\pm\infty$ values.

## 1.8. Measurable Neutrosophic Set and Measurable Neutrosophic Space.

The members of $\Sigma$ are called measurable neutrosophic sets, while $(X,\Sigma)$ is called a measurable neutrosophic space.

### 1.9. Neutrosophic Measurable Function.

A function $f:(X,\Sigma_X)\to(Y,\Sigma_Y)$, mapping two measurable neutrosophic spaces, is called *neutrosophic measurable function* if $\forall B\in\Sigma_Y, f^{-1}(B)\in\Sigma_X$ (the inverse image of a neutrosophic $Y$-measurable set is a neutrosophic $X$-measurable set).

### 1.10. Neutrosophic Probability Measure.

As a particular case of neutrosophic measure $\nu$ is the neutrosophic probability measure, i.e. a neutrosophic measure that measures probable/possible propositions

$$^{-}0\leq\nu(X)\leq 3^{+}, \qquad (5)$$

where $X$ is the whole neutrosophic probability sample space.

We use nonstandard numbers, such $1^{+}$ for example, to denominate the absolute measure (measure in all possible worlds), and standard numbers such as *1* to denominate the relative measure (measure in at least one world). Etc.

We denote the neutrosophic probability measure by $\mathcal{NP}$ for a closer connection with the classical probability $\mathcal{P}$.

### 1.11. Neutrosophic Category Theory.

The neutrosophic measurable functions and their neutrosophic measurable spaces form a neutrosophic category, where the functions are arrows and the spaces objects.

We introduce the neutrosophic category theory, which means the study of the neutrosophic structures and of the neutrosophic mappings that preserve these structures.

The classical category theory was introduced about *1940* by Eilenberg and Mac Lane.

A neutrosophic category is formed by a class of neutrosophic objects $X,Y,Z,...$ and a class of neutrosophic morphisms (arrows) $\nu,\xi,\omega,...$ such that:

a) If $Hom(X,Y)$ represent the neutrosophic morphisms from $X$ to $Y$, then $Hom(X,Y)$ and $Hom(X',Y')$ are disjoint, except when $X=X'$ and $Y=Y'$;
b) The composition of the neutrosophic morphisms verify the axioms of
   i) Associativity: $(\nu\circ\xi)\circ\omega=\nu\circ(\xi\circ\omega)$
   ii) Identity unit: for each neutrosophic object $X$ there exists a neutrosophic morphism denoted $id_X$, called *neutrosophic identity* of $X$ such that $id_X\circ\nu=\nu$ and $\xi\circ id_X=\xi$

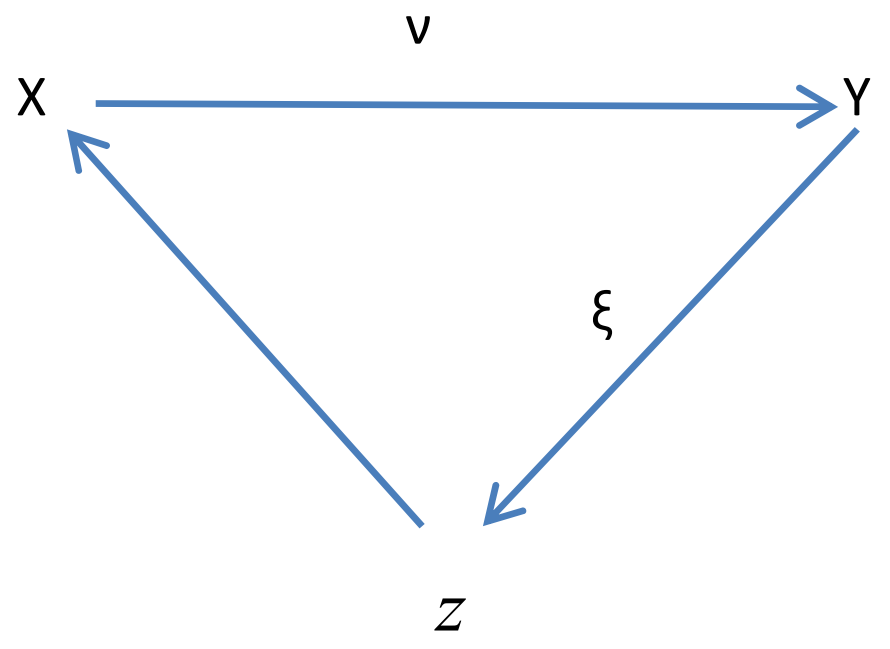

*Fig. 2*

## 1.12. Properties of Neutrosophic Measure.

*a) Monotonicity.*

If $A_1$ and $A_2$ are neutrosophically measurable, with $A_1 \subseteq A_2$, where

$$\nu(A_1) = \left(m(A_1), m(neutA_1), m(antiA_1)\right),$$

and

$$\nu(A_2) = \left(m(A_2), m(neutA_2), m(antiA_2)\right),$$

then

$$m(A_1) \le m(A_2), m(neutA_1) \le m(neutA_2), m(antiA_1) \ge m(antiA_2). \quad (6)$$

Let $\nu(X) = (x_1, x_2, x_3)$ and $\nu(Y) = (y_1, y_2, y_3)$. We say that $\nu(X) \le \nu(Y)$, if $x_1 \le y_1, x_2 \le y_2$, and $x_3 \ge y_3$.

*b) Additivity.*

If $A_1 \cap A_2 = \Phi$, then $\nu(A_1 \cup A_2) = \nu(A_1) + \nu(A_2)$, (7)

where we define

$$(a_1, b_1, c_1) + (a_2, b_2, c_2) = (a_1 + a_2, b_1 + b_2, a_3 + b_3 - m(X)), \quad (8)$$

where $X$ is the whole neutrosophic space, and

$$a_3 + b_3 - m(X) = m(X) - m(A) - m(B) = m(X) - a_1 - a_2$$
$$= m(antiA \cap antiB).$$
(9)

### 1.13. Neutrosophic Measure Continuous from Below or Above.

A *neutrosophic measure* $\nu$ *is continuous from below* if, for $A_1, A_2, ...$ neutrosophically measurable sets with $A_n \subseteq A_{n+1}$ for all $n$, the union of the sets $A_n$ is neutrosophically measurable, and

$$\nu\left(\bigcup_{n=1}^{\infty} A_n\right) = \lim_{n\to\infty} \nu(A_n) \tag{10}$$

And a *neutrosophic measure* $\nu$ *is continuous from above* if for $A_1, A_2, ...$ neutrosophically measurable sets, with $A_n \supseteq A_{n+1}$ for all $n$, and at least one $A_n$ has finite neutrosophic measure, the intersection of the sets $A_n$ and neutrosophically measurable, and

$$\nu\left(\bigcap_{n=1}^{\infty} A_n\right) = \lim_{n\to\infty} \nu(A_n). \tag{11}$$

### 1.14. Generalizations.

1.14.a. Neutrosophic measure is a generalization of the *fuzzy measure*, because when $m(neutA) = 0$ and *m(antiA)* is ignored, we get

$$\nu(A) = (m(A), 0, 0) \equiv m(A), \tag{12}$$

and the two fuzzy measure axioms are verified:

a) If $A = \Phi$, then $\nu(A) = (0,0,0) \equiv 0$

b) If $A \subseteq B$, then $\nu(A) \leq \nu(B)$.

1.14.b. The neutrosophic measure is practically a triple classical measure: a classical measure of the determinate part of a neutrosophic object, a classical part of the indeterminate part of the neutrosophic object, and another classical measure of the determinate part of the opposite neutrosophic object. Of course, if the indeterminate part does not exist (its measure is zero) and the measure of the opposite object is ignored, the neutrosophic measure is reduced to the classical measure.

### 1.15. Examples.

Let's see some examples of neutrosophic objects and neutrosophic measures.

a) If a book of *100* sheets (covers included) has *3* missing sheets, then

$$\nu(book) = (97,3,0), \tag{13}$$

where $\nu$ is the neutrosophic measure of the book number of pages.

b) If a surface of *5* × *5* square meters has cracks of *0.1* × *0.2* square meters, then

$$\nu(surface) = (24.98,0.02,0), \tag{14}$$

where $\nu$ is the neutrosophic measure of the surface.

c) If a die has two erased faces then

$$v(die) = (4, 2, 0),$$

where $v$ is the neutrosophic measure of the die's number of correct faces.

d) An approximate number $N$ can be interpreted as a neutrosophic measure $N = \underline{d} + \underline{i}$, where $\underline{d}$ is its determinate part, and $\underline{i}$ its indeterminate part. Its anti part is considered *0.*

For example if we don't know exactly a quantity $q$, but only that it is between let's say $q \in [0.8, 0.9]$, then $q = 0.8 + i$, where *0.8* is the determinate part of $q$, and its indeterminate part $i \in [0, 0.1]$.

We get a negative neutrosophic measure if we approximate a quantity measured in an inverse direction on the *x*-axis to an equivalent positive quantity.

For example, if $r \in [-6, -4]$, then $r = -6 + i$, where -6 is the determinate part of r, and $i \in [0, 2]$ is its indeterminate part. Its anti part is also 0.

e) Let's measure the truth-value of the proposition
*G* = "through a point exterior to a line one can draw only one parallel to the given line".
The proposition is incomplete, since it does not specify the type of geometrical space it belongs to. In an *Euclidean geometric space* the proposition *G* is true; in a *Riemannian geometric space* the proposition *G* is false (since there is no parallel passing through an exterior point to a given line); in a *Smarandache geometric space* (constructed from mixed spaces, for example from a part of Euclidean subspace together with another part of Riemannian space) the proposition $G$ is indeterminate (true and false in the same time).

$$\nu(G) = (1,1,1). \tag{15}$$

f) In general, not well determined objects, notions, ideas, etc. can become subject to the neutrosophic theory.

# Chapter 2:
# Introduction to Neutrosophic Integral

## 2.1. Definition of Neutrosophic Integral

Using the neutrosophic measure, we can define a neutrosophic integral.
The neutrosophic integral of a function $f$ is written as:

$$\int_X f d\nu \qquad (16)$$

where $X$ is the a neutrosophic measure space,
and the integral is taken with respect to the neutrosophic measure $\nu$.

Indeterminacy related to integration can occur in multiple ways: with respect to value of the function to be integrated, or with respect to the lower or upper limit of integration, or with respect to the space and its measure.

## 2.2. *First Example of Neutrosophic Integral: Indeterminacy Related to Function's Values*

Let

$$f_N: [a, b] \to R \qquad (17)$$

where the neutrosophic function is defined as:

$$f_N(x) = g(x)+i(x) \qquad (18)$$

with $g(x)$ the determinate part of $f_N(x)$, and $i(x)$ the indeterminate part of $f_N(x)$, where for all $x$ in $[a, b]$ one has:

$$i(x) \in [0, h(x)], h(x) \geq 0 . \qquad (19)$$

Therefore the values of the function $f_N(x)$ are approximate, i.e.

$$f_N(x) \in [g(x), g(x) + h(x)] . \qquad (20)$$

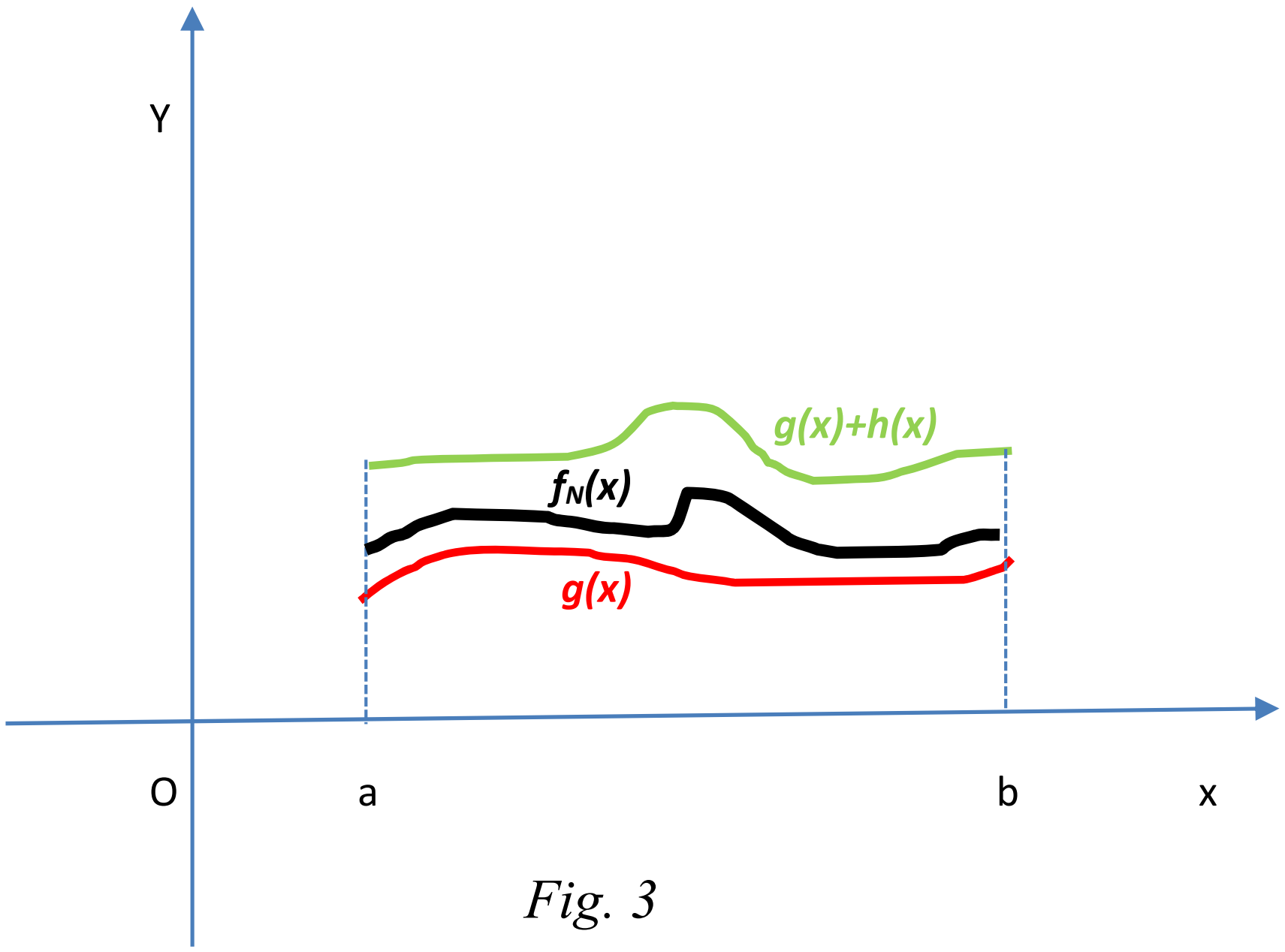

*Fig. 3*

Similarly, the neutrosophic integral is an approximation:

$$\int_a^b f_N(x)dv = \int_a^b g(x)dx + \int_a^b i(x)dx \qquad (21)$$

## *2.3. Second Example of Neutrosophic Integral: Indeterminacy Related to the Lower Limit*

Suppose we need to integrate the function

$$f: X \to R \qquad (22)$$

on the interval *[a, b]* from *X*, but we are unsure about the lower limit *a*. Let's suppose that the lower limit "*a*" has a determinant part "$a_1$" and an indeterminate part *ε*, i.e.

$$a = a_1 + \varepsilon \qquad (23)$$

where

$$\varepsilon \in [0, 0.1]. \qquad (24)$$

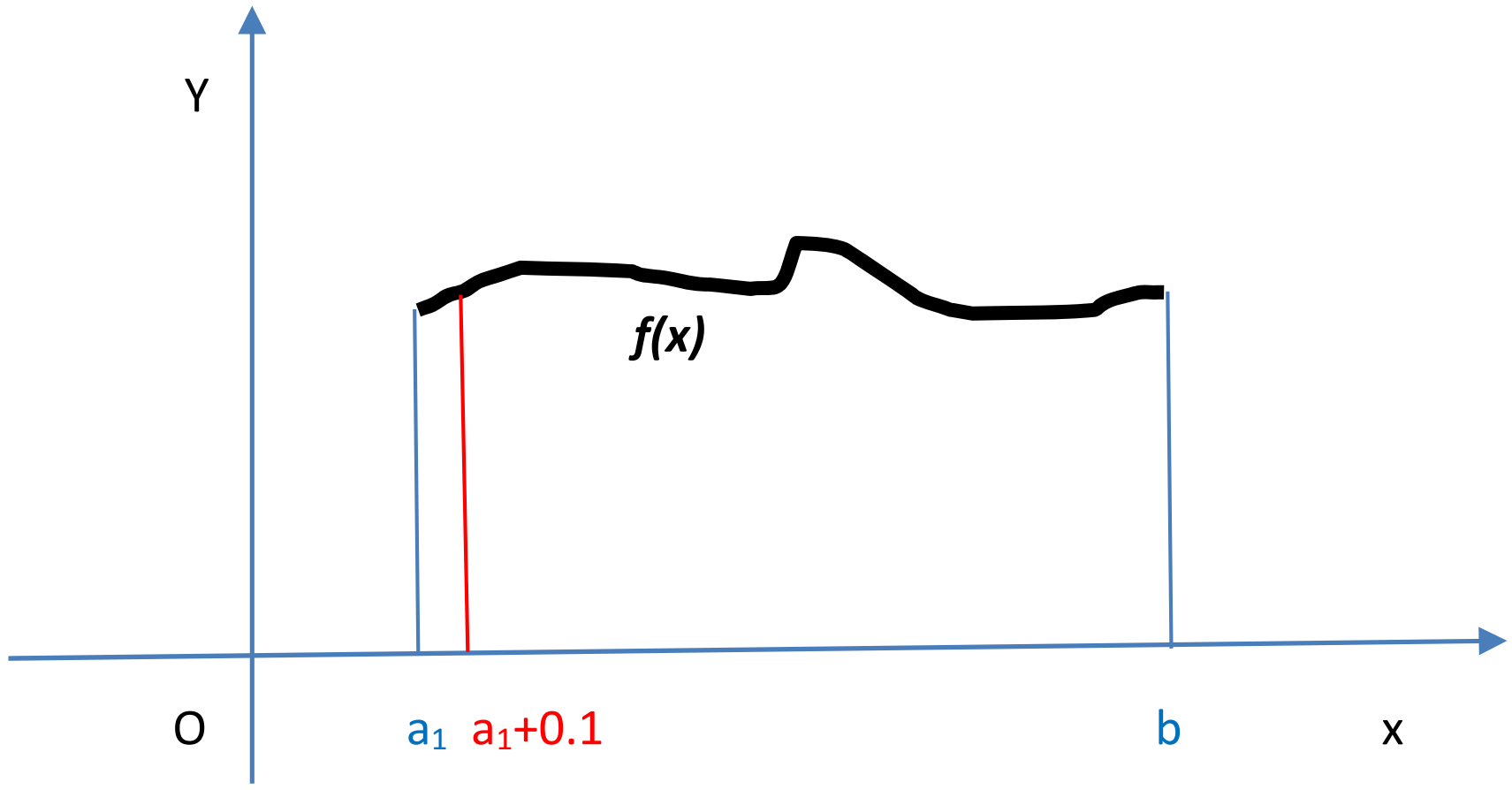

*Fig. 4*

Therefore

$$\int_{a}^{b} {}_X f d\nu = \int_{a_1}^{b} f(x)dx - \mathrm{i}_1 \tag{25}$$

where the indeterminacy $i_1$ belongs to the interval:

$$i_1 \in [0, \int_{a_1}^{a_1+0.1} f(x)dx]. \tag{26}$$

Or, in a different way:

$$\int_{a}^{b} {}_X f d\nu = \int_{a_1+0.1}^{b} f(x)dx + \mathrm{i}_2 \tag{27}$$

where similarly the indeterminacy $i_2$ belongs to the interval:

$$i_2 \in [0, \int_{a_1}^{a_1+0.1} f(x)dx]. \tag{28}$$

# Chapter 3:
# Introduction to Neutrosophic Probability

### 3.1. First Example of Indeterminacy.

The idea of extending the neutrosophic principle, which is based on indeterminacy, to probability, came to my mind when I tossed a die outside, on my stairs made of concrete, but the concrete was broken, had small cracks, and the die got stuck on an edge in a crack. There was no clear face to see, hence it was an indeterminacy.

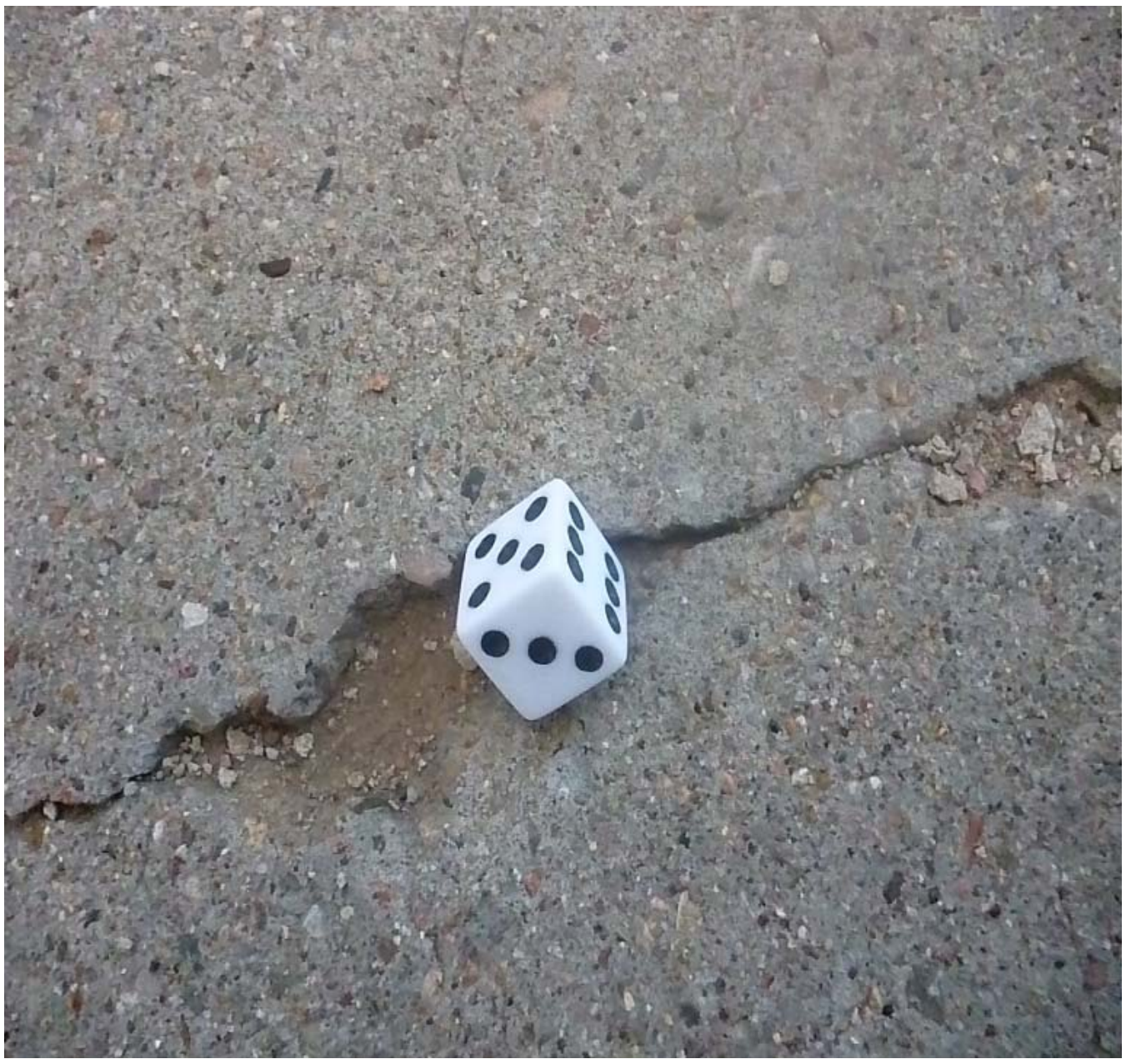

*Fig. 5. Indeterminate die state*

Thus tossing a die on a cracked surface one can get:

$$\{1, 2, 3, 4, 5, 6, \ indeterminacy\}.$$

This is its sample space.

A cubic die (with 12 edges and 8 vertices) tossed on an irregular surface has the chance to fall on a vertex or on an edge in a small slit or crack (not on one of its faces). Therefore, tossing the die can turn on an indeterminate outcome.

Whence, the neutrosophic probability $\mathcal{NP}_T$ of tossing, for example $\{1\}$ is less than $\frac{1}{6}$, since there are seven possible outcomes $\mathcal{NP}_T(1) < \frac{1}{6}$, not like in classical probability where $\mathcal{P}(1) = \frac{1}{6}$.

The more irregularities on the surface (as below), the more indeterminacy occurs:

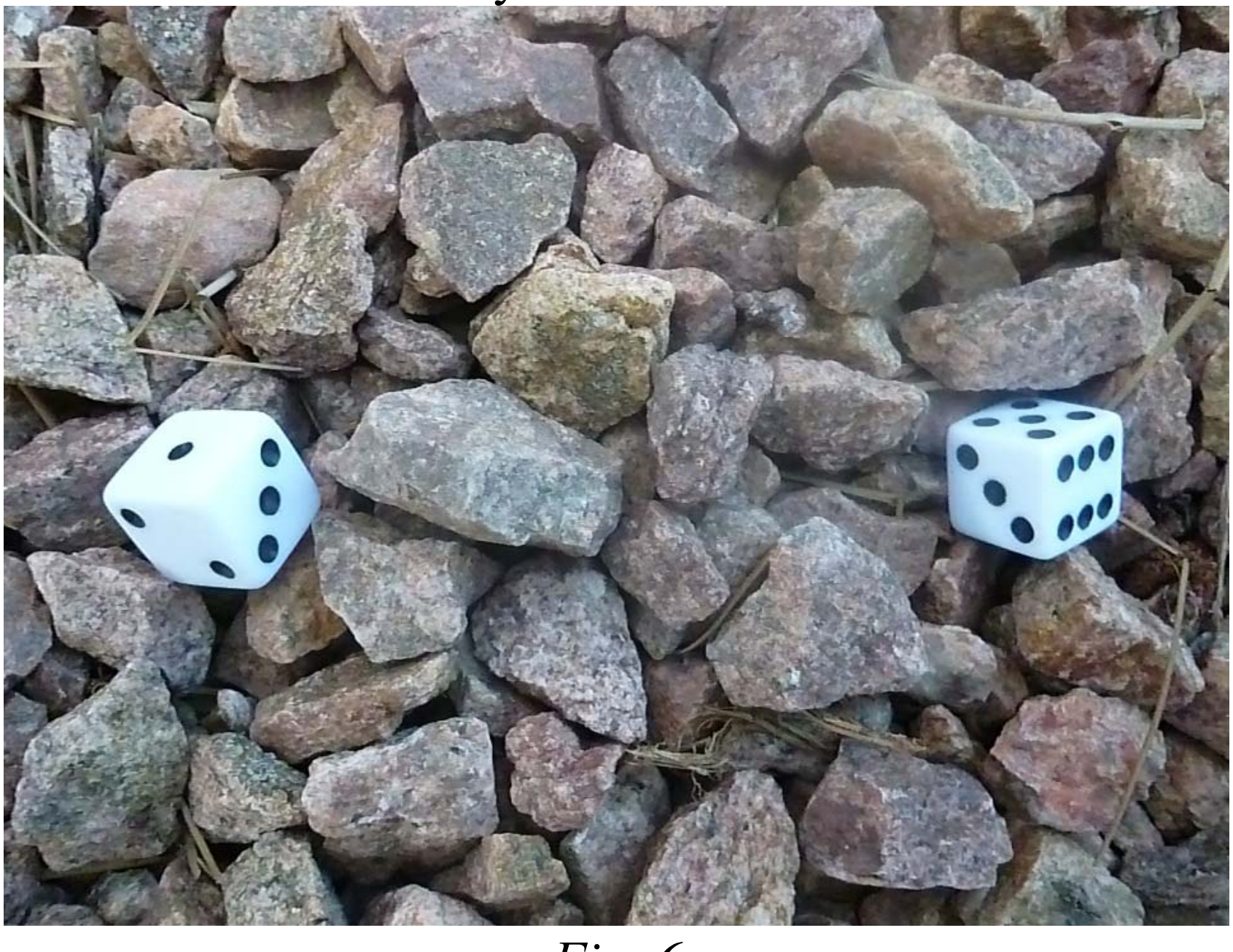

*Fig. 6*

In the classical probability the die and the surface it rolls on are considered perfect, hence there is no indeterminacy due to the materials. There is only randomness.

In neutrosophic probability one has, besides randomness, indeterminacy due to construction materials and shapes of the die and of the surface.

If the die is not regular, and the faces have different areas, or the die's center of mass is not in the geometrical die's center, then the probability will be proportional to the face's surface, and the closer is the center of mass to a face the higher the probability for that face.

The die's mass of inhomogeneous density will influence the probability outcome.

### 3.2. Second Example of Indeterminacy.

Let's consider a regular die (with six faces), having two faces whose print is erased (let's say faces 5 and 6). Then:

$$\mathcal{NP}(1) = \mathcal{NP}(2) = \mathcal{NP}(3) = \mathcal{NP}(4) = \frac{1}{6},$$

$$\mathcal{NP}(5) = \mathcal{NP}(6) = 0,$$

while $\mathcal{NP}(indeterm) = \dfrac{2}{6}$, (29)

when the die is tossed on a regular surface.

### 3.3. Third Example of Indeterminacy.

On a surface with cracks there is a chance that flipping a coin, the coin falls into a crack and gets stuck on its edge; then we have again indeterminacy.

$$\mathcal{NP}(Head) = \mathcal{NP}(Tale) < ½ \quad (30)$$

and the sample space is {Head, Tale, indeterminacy}.

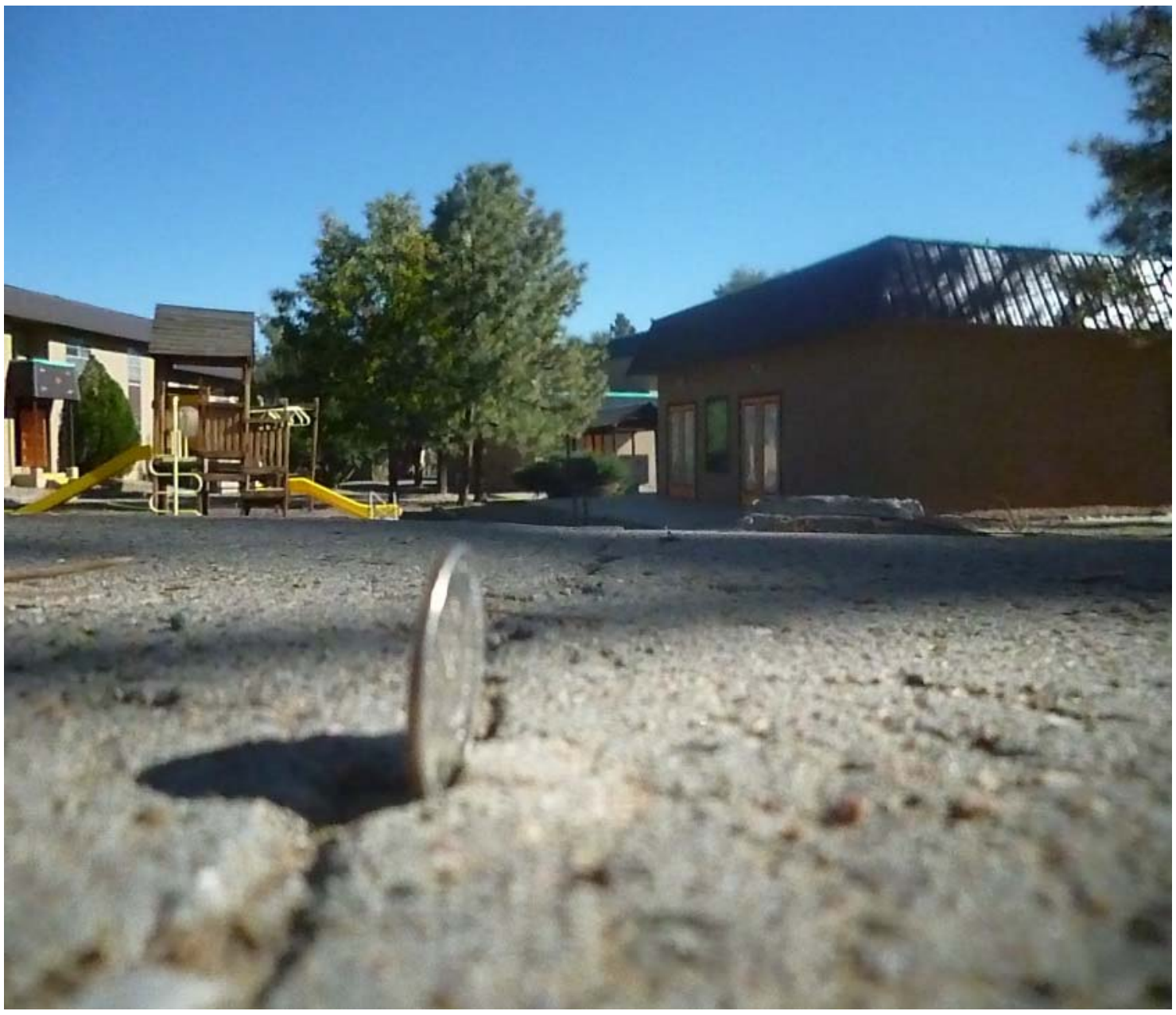

*Fig. 7. Indeterminacy related to tossing a coin*

### 3.4. Fourth Example of Indeterminacy.

An urn with two types of votes: *A*-ballots and *B*-ballots, but some votes are deteriorated, and we can't determine if it's written *A* or *B*. Therefore, we have indeterminate votes.

In many practical applications we may not even know the exact number of indeterminate votes, or of *A*-ballots, or of *B*-ballots. Therefore, the indeterminacy is even bigger.

### 3.5. Fifth Example of Indeterminacy.

If there are two candidates *A* and *B* for presidency, and the probability that *A* wins is *0.46*, it doesn't mean that the probability that *B* wins is *0.54*, since there may be blank votes (from the voters not choosing any candidate) or black votes (from the voters that reject both candidates).

For example, the probability that *B* wins could be *0.45*, while the difference *1-0.46-0.45 = 0.09* would be the probability of blank and black votes together. Therefore we have a neutrosophic probability: $\mathcal{NP}(A) = (0.46, 0.09, 0.045)$

### 3.6. Sixth Example of Indeterminacy.

If a meteorology center reports that the chance of rain tomorrow is 60%, it does not mean that the chance of not raining is *40%*, since there might be hidden parameters (weather factors) that the meteorology center is not aware of.

There might be an unclear weather, for example, cloudy and humid day, that some people can interpret as rainy day and others as non-rainy day. The ambiguity arouses indeterminacy.

### 3.7. The Seventh Example of Indeterminacy.

If some drug tests are *95%* reliable, it doesn't mean that *5%* they are unreliable, because there might be some unknown effects of the drugs that we are not sure they are beneficial or harmful.

**3.8. The Eighth Example of Indeterminacy.**

A roulette wheel has *38* numbers. But, having been used too much, several of its numbers have been erased, and one cannot read.

Therefore, we get again indeterminacy.

**3.9. Ninth Example of Indeterminacy.**

A deck of *52* cards has *3* damaged cards that we are unable to read. Then we have indeterminacy. If the damaged cards are visibly broken, we don't have equiprobability.

**3.10. Tenth Example of Indeterminacy.**

Probability in a soccer game.

Classical probability is incomplete, because it computes for a team the chance of winning, or the chance of not winning, nut not all three chances as in neutrosophic probability: winning, having tie game, or losing.

**3.11. Eleventh Natural Example of Indeterminacy.**

Indeterminacy occurs (yet rarely) whether a series of newborns will be girls or boys, since some transgender children can have undetermined (ambiguous) sex, i.e. partially male and partially female).

## 3.12. Example of a Neutrosophic Continuous Random Variable.

The previous examples used neutrosophic discrete random variables.

Let’s now consider a spinner as bellow:

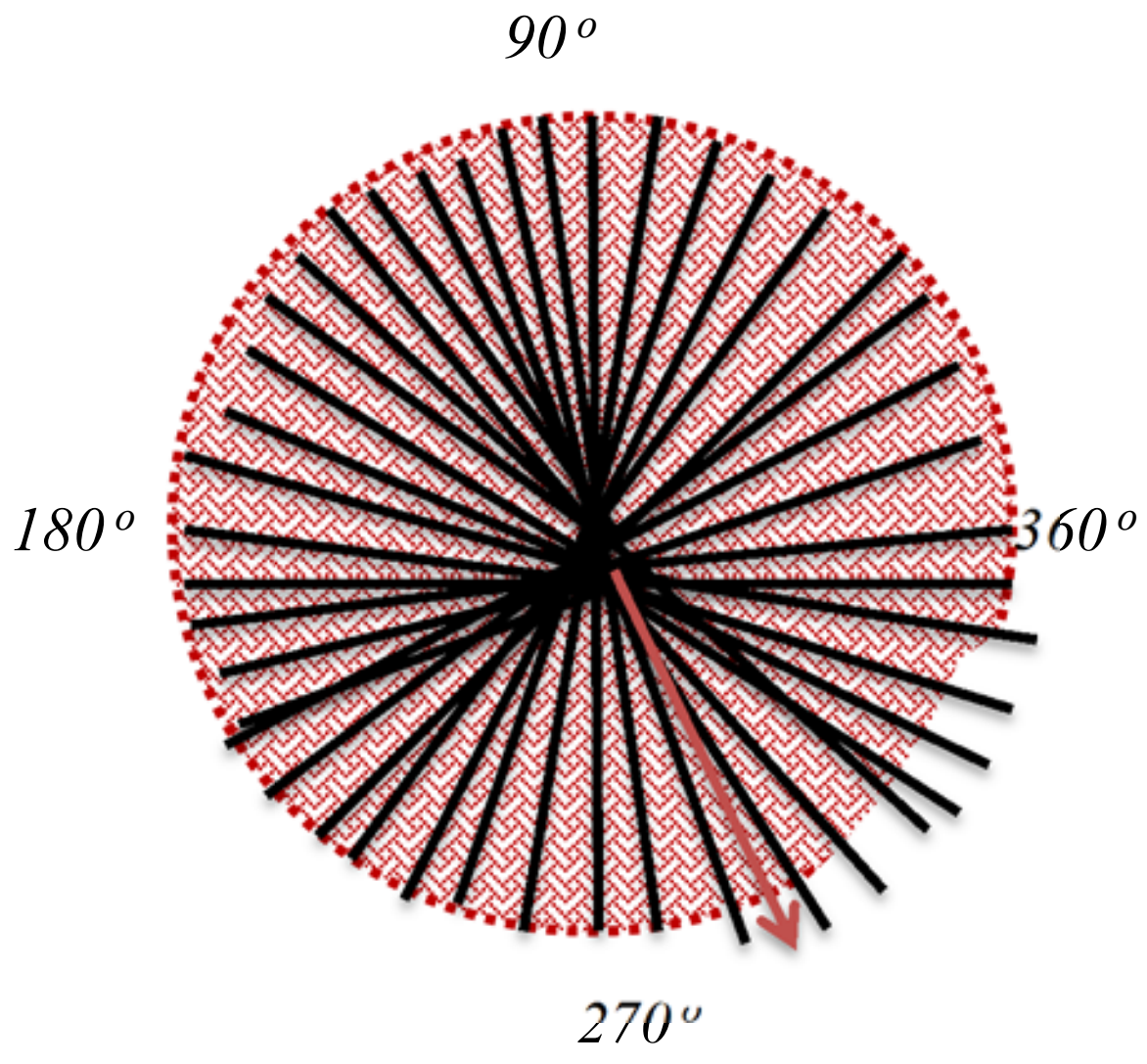

*Fig. 8*

The continuous sample space is $\Omega = [0,360]$. Let’s say that spinner’s table is erased between *270°* – *360°*, so if the spinner gets in this area ($IV^{th}$ quadrant) we are not able to read a number, we consider it indeterminacy zone. Therefore $\mathcal{NP}(indeterm)=\frac{1}{4}$. We have a continuous random variable.

$$\mathcal{NP}([90,100])=\left(ch([90,100]),ch(indeterm),ch\left(\overline{[90,100]}\right)\right)$$
$$=\left(\frac{10}{360},\frac{90}{360},\frac{260}{360}\right) \quad (31)$$

### 3.13. First Types of Indeterminacies.

One has at least two types of indeterminacies:

a) The indeterminacy due to the space (for example the surface on which the dices are tossed on, the urn on which the votes are introduced, etc.).

b) The indeterminacy due to the items contained into the physical space (for example the defect dice, the unclear ballots, etc.).

### 3.14. Second Types of Indeterminacies.

a) We have indeterminacy not related to a particular event, which is a *constant indeterminacy*. For example, tossing a regular die on a irregular surface which has cracks. No matter what outcome we look for 1, 2, ..., or 6, the indeterminacy (chance that the die falls in a crack and has an unclear reading) is the same.

b) But we may have *indeterminacy related to each event*. For example, let the sample space be:

$$\{sunny\ day, rainy\ day, snowfall\ day\}$$

as a weather forecast for one weak from today. A metheorologist approximately computes the chance of each event, using various parameters, such as: statistics of past weather, today's weather, etc. and gives the following (imprecise) probabilities:

$\{[0.1, 0.2], [0.5, 0.7], [0.3, 0.6]\}$, (32)

where [0.1, 0.2] means the probability of sunny day,

[0.5, 0.7] probability of rainy day,

and [0.3, 0.6] probability of snowfall day.

Thus, we have different indeterminacies which are related to the occurrence of each event.

Neutrosophically, we can write it as:

$NP(sunny\ day) = 0.1 + i_1$, where $i_1 \in [0.0, 0.1]$,

$NP(rainy\ day) = 0.5 + i_2$, where $i_2 \in [0.0, 0.2]$,

and

$NP(snowfall\ day) = 0.3 + i_3$, where $i_3 \in [0.0, 0.3]$

with $i_1, i_2, i_3$ indeterminacies.

Let's compute the union of events

*NP*(*sunny day* **or** *snowfall day*).

$$NP(sunny\ day) + NP(snowfall\ day) = (0.1 + i_1) + (0.3 + i_3) = (0.1 + 0.3) + (i_1 + i_3) = 0.4 + i_4,$$

where $i_4 \in [0.0, 0.1] + [0.0, 0.3] = [0.0, 0.4]$.

This could also be computed simply as in classical imprecise probability:

$P(sunny\ day$ **or** $snowfall\ day) =$ $[0.1, 0.2] + [0.3, 0.6] = [0.4, 0.8] = 0.4 + i_4$,

where $i_4 \in [0.0, 0.4]$.

Similarly for intersection of events:

$$NP\ (sunny\ \textbf{and}\ snowfall\ day) = (0.1 + i_1) \cdot (0.3 + i_3) = (0.1)(0.3) + \{0.3i_1 + 0.1i_3 + i_1 i_3\} = 0.03 + \{[0.0, 0.3] + [0.0, 0.3] + [0.0, 0.3]\} = 0.03 + i_5,$$

where $i_5 \in [0.0, 0.9]$.

This is because:

$$\{Indeterminacy\} \cdot \{number\} = \{indeterminacy\}$$

and

$$\{indeterminacy\} \cdot \{indeterminacy\} = \{indeterminacy\}.$$

Classically:

$$P(sunny\ day\ \textbf{and}\ snowfall\ day) =$$
$$[0.1, 0.2] \cdot [0.3, 0.6] = [0.03, 0.12] = 0.03 + i_5,$$
where $i_5 \in [0.0, 0.09]$.

Similarly for negation of events:

$$NP(\textbf{not}\ a\ sunny\ day) = 1 - (0.1 + i_1)$$
$$= 1 - 0.1 - i_1 = 0.9 - i_1 =$$
$$= 0.8 + i_6, where\ i_6 \in [0.0, 0.1].$$

Classically:

$$P(\textbf{not}\ a\ sunny\ day) = 1 - [0.1, 0.2]$$
$$= [0.8, 0.9] = 0.8 - i_6,$$

where $i_6 \in [0.0, 0.1]$.

c) Or *mixt indeterminacies*: to some events there is a chance of indeterminacy > 0, while to other events there is not.

A similar example as the previous, but we change the data:

$$\{[0.1, 0.2], [0.5, 0.7], 0.3\}. \quad (33)$$

Therefore, there is indeterminacy related to the first and second events, but not to the third.

### 3.15. Distinction between Indeterminacy and Randomness.

Indeterminacy is different from randomness. Indeterminacy is due to the defects of the construction of the physical space (where an event can occur), and/or to the imperfect construction of the physical objects involved in the event, etc.

Therefore, neutrosophic probability analyses both: the random phenomena, and the indeterminacy related to these phenomena.

In consequence, neutrosophic probability deals with two types of variables: random variables and indeterminacy variables, and two types of processes: stochastic process and respectively indeterminate process.

### 3.16. Neutrosophic Random Variables.

A *classical random (stochastic) variable* is subject to change due to randomness, while the *neutrosophic random (stochastic) variable* is subject to change due to both randomness and indeterminacy.

A neutrosophic random variable's values represent the possible outcomes and possible indeterminacies. The randomness and indeterminacy can be objective or subjective.

Alike classical random variables, the neutrosophic random variables can be classified as:

- *discrete*, that is it can take a value in a specified list of exact values and a finite number of indeterminacies;
- *continuous*, that is it can take a value or an indeterminacy in an interval, or in a collection of intervals;

- *mixt*, that is it can take a value or indeterminacy either in a specified list of exact values, or in an interval or in a collection of intervals (mixture of discrete and continuous).

Another classification, alike classical random variables, for neutrosophic random variables is

- *finite*; having of course a finite number of possible outcomes and possible indeterminacies;
- *infinite*; having an infinite number of possible outcomes or indeterminacies.

An infinite neutrosophic random variable can be

- *countably*;
- or *uncountably*.

A *neutrosophic random variable* $X$ is *admissible* if it is possible to compute the chance that the value of $X$ is less than any particular number, together with its corresponding indeterminacy and its nonchance. Which is equivalent to the possibility of computing the chance that the value of $X$ is in any range, range that must be mapped to a subset of the neutrosophic sample space $\nu\Omega$.

### 3.17. Many Possible Neutrosophic Measures and Probabilities.

We may be able to define the neutrosophic measure and neutrosophic probability in many ways, since we work with approximations and indeterminacies. Their definitions may depend on each particular application.

### 3.18. Definition of Neutrosophic Probability

Neutrosophic probability (or likelihood) is a particular case of the neutrosophic measure. It is an estimation of an event (different from indeterminacy) to occur, together with an estimation that some indeterminacy may occur, and the estimation that the event does not occur.

Neutrosophic Probability and Neutrosophic Statistics started in *1995*, but was not developed and applied as much as neutrosophic logic and neutrosophic set that are widely used.

A **neutrosophic random variable** is a variable that may have an indeterminate (unclear, ambiguous) outcome.

A **neutrosophic random (stochastic) process** represents the evolution over time of some neutrosophic random values. It is a collection of neutrosophic random variables.

The classical probability deals with fair dice, coins, roulettes, spinners, decks of cards, random walks, while neutrosophic probability deals with unfair, imperfect such objects, variables and processes.

The neutrosophic probability is a generalization of the classical probability because, when the chance of indeterminacy of a stochastic process is zero, these two probabilities coincide.

### 3.19. Neutrosophic Probability vs. Imprecise Probability.

In Imprecise Probability ($\mathcal{IP}$), the probability of an event $A$,

$$\mathcal{IP}(A) = (a,b) \subseteq [0,1] \tag{34}$$

is an interval included into $[0,1]$, not a crisp number.

The Neutrosophic Probability that an event $A$ occurs is

$$\mathcal{NP}(A) = \big(ch(A), ch(neutA), ch(antiA)\big) = (T, I, F), \tag{35}$$

but sometimes instead of "neutA" we say "indeterminacy related to A" and we denote it by "indeterm$_A$"; also we note "antiA" by $\overline{A}$;
where $T, I, F$ are standard or nonstandard subsets of the nonstandard unitary interval $]^-0, 1^+[$, and $T$ is the chance that $A$ occurs, denoted *ch(A)*; $I$ is the indeterminate chance related to A, *ch(indeterm$_A$);* and $F$ is the chance that $A$ does not occur, $ch(\overline{A})$.

So, $\mathcal{NP}$ is a generalization of the Imprecise Probability as well.

Therefore, using other notations we have:

$$\mathcal{NP}(A) = \big(ch(A), ch(indeterm_A), ch(\overline{A})\big). \tag{36}$$

We used the notations $T$ (truth), $I$ (indeterminate), and $F$ (falsehood) in order to be consistent with those from neutrosophic logic and neutrosophic set, widely spread.

In the most general case, $T,I,F$ are standard or non-standard subsets of the unitary non-standard interval $\left]^{-}0,1^{+}\right[$, in order to be able to make distinction between *absolute sure event* (sure event in all possible worlds -- whose probability value is $1^{+}$), and *relative sure event* (i.e. sure event in at least one world, but not in all words -- whose probability is $1$, where $1<1^{+}$).
Similarly, for *absolute impossible event* (impossible event in all possible worlds -- whose probability is $^{-}0$), and *relative impossible event* (i.e. impossible event in at least one world, but not in all words -- whose probability is $^{-}0$, where $^{-}0<0$).

$$1^{+}=1+\varepsilon \text{ and } ^{-}0=0-\varepsilon, \quad (37)$$

where $\varepsilon$ is a very tiny positive number.

For technical applications we'll use only standard sets and the standard unit interval $[0,1]$. And throughout this book, with few exceptions.

Let's note by majuscules the subsets $T,I,F$ and by lower-case letters the crisp numbers $t,i,f$. For the crisp neutrosophic probability, when $T,I,F$ are just standard or non-standard numbers in $\left]^{-}0,1^{+}\right[$, in the most general case one has:

$$^{-}0\leq t+i+f\leq 3^{+}, \quad (38)$$

considering that the tree components $t,i,f$ are independent (as in neutrosophic logic and in neutrosophic set).

If only two components are dependent, while the third one is independent from them, then

$$^{-}0 \leq t+i+f \leq 2^{+}. \quad (39)$$

If all three components are dependent two by two, then

$$^{-}0 \leq t+i+f \leq 1^{+}. \quad (40)$$

Let's consider the standard case.

1) If $t+i+f=1$ one has *complete probability* (the most common application), or *normalized probability*.
2) If $t+i+f<1$ one has *incomplete probability* (because the source of information or the stochastic process is incomplete, i.e. not well known).
3) If $t+i+f>1$ one has *paraconsistent probability* (because of conflicting sources of information that transmit us contradictory information; for example one source may compute the chance that an event occurs using some criteria (parameters influencing the event), but it is not able to compute the chance that the event does not occur, while another independent source of information may compute the chance that the event does not occur using different criteria (different parameters), but not able to compute the chance that the event occur.

Similarly, for computing the chance of indeterminacy of the stochastic process by a third independent source of information. Therefore it is possible to get the sum $t+i+f\neq 1$. (41)

### 3.20. Sigma-Algebra of Events.

A sigma-algebra or $\sigma$-algebra of $X$, in the measure theory, is a collection of subsets of the set $X$ such that

1) $\Phi\in\Sigma$;
2) $X\in\Sigma$;
3) If $A\in\Sigma$ then the complement of $A$, $C(A)\in\Sigma$;
4) If $A_1, A_2, ..., A_n\in\Sigma$, then the countable union $A_1\cup A_2\cup,...,\cup A_n\in\Sigma$.

### 3.21. Definition of Classical Probability.

The classical probability measure is a mapping:

$$\mathcal{P}: X\to[0,1] \tag{42}$$

where $X$ is a sample space, such that $\mathcal{P}(X)=1$ and $\mathcal{P}$ is additive for the union

$$\mathcal{P}(A\cup B)=\mathcal{P}(A)+\mathcal{P}(B) \text{ for } A\cap B=\phi, \tag{43}$$

even for infinite unions:

$$\mathcal{P}\left(\bigcup_{n\geq 0}A_i\right)=\sum_{n\geq 0}\mathcal{P}(A_n) \tag{44}$$

for $A_i$ disjoint two by two, that lie in the sigma-algebra of events $\Sigma$ of $X$.

### 3.22. Neutrosophic Sigma-Algebra of Events.

The neutrosophic sigma-algebra of events $\nu\Sigma$ will be defined in the same way, with the distinction that the set $X$ contains some indeterminacy. Therefore there are some subjects of $X$ that are indeterminate parts.

### 3.23. Neutrosophic Probability Measure.

The neutrosophic probability measure is a mapping:

$$\mathcal{NP} : X \to [0,1]^3 \tag{45}$$

where $X$ is a neutrosophic sample space (i.e. $X$ contains some indeterminacy),

$$\mathcal{NP}(A) = \left(ch(A), ch(indeterm_A), ch\left(\overline{A}\right)\right), \tag{46}$$

or, using other notations, we have:

$$\mathcal{NP}(A) = \left(ch(A), ch(neutA), ch(antiA)\right) \tag{47}$$

where $indeterm_A$ means the indeterminacy that may occur when trying to have event $A$ occurs,

such that the neutrosophic probability of the whole space $X$ has the property that:

$$\mathcal{NP}(X) = (\alpha, \beta, \gamma),$$

where ${}^-0 \le \alpha, \beta, \gamma \le 1^+$, and

$$^-0 \le \alpha + \beta + \gamma \le 3^+. \tag{48}$$

Therefore, the sum of the three components of the neutrosophic probability of the whole sample space is not required to be equal to *1* as in classical probability, since there cases where it is strictly less than *1*, or strictly greater than *1*.

We also have:

$$\mathcal{NP}(A \cup B) = \left(ch(A) + ch(B), ch(indeterm_{A \cup B}), ch\left(\overline{A \cup B}\right)\right) \tag{49}$$

for $A \cap B = \phi$, and for infinite unions:

$$\mathcal{NP}\left(\bigcup_{n \geq 0} A_n\right) = \left(\sum_{n \geq 0} ch(A_n), ch\left(indeterm_{\underset{n \geq 0}{\cup} A_n}\right), ch\left(\overline{\bigcup_{n \geq 0} A_n}\right)\right) \tag{50}$$

for $A_n$ disjoint two by two that lie in the neutrosophic sigma algebra of events.

*Remark.* Although in most cases the sum of the three components is *1* (in normalized probability):

*ch(A) + ch(neutA) + ch(antiA) = 1* *(51)*

or using similar notations

$ch(A) + ch(indeterm_A) + ch\left(\overline{A}\right) = 1,$ *(52)*

we still recommend to computing all three components because it arises cases when the probability is not normalized.

## 3.24. Neutrosophic Probability Mass Function.

A *Neutrosophic Probability Mass Function* $(npmf)$ is a function

$$f: \nu\Omega \longrightarrow [0,1]^3$$

$$f(x) \in [0,1]^3 \text{ for all } x \in \nu\Omega,$$

$$f(x) = (\ ch(x), ch(indeterm_x), ch(\bar{x})\ )\ ). \tag{53}$$

A *Neutrosophic Event* is any subset $E$ of the neutrosophic sample space $\nu\Omega$. The neutrosophic probability of any event $E$ is defined as:

$$NP(E) = \left(\sum_{x \in E} ch(x), ch(indeterm_E), \sum_{y \in \bar{E}} ch(y)\right). \quad (54)$$

### 3.25. Neutrosophic Probability Axioms.

They are extensions of Kolmogorov axioms from classical probability.

$(\nu\Omega, NF, NP)$ is a neutrosophic probability space, where $\nu\Omega$ is a neutrosophic sample space, *NF* is a neutrosophic event space, and *NP* is a neutrosophic probability measure.

*First Axiom.*

The neutrosophic probability of an event $A$

$$NP(A) = (\, ch(A), ch(indeterm_A), ch(\bar{A})\,), \quad (55)$$

where $ch(A) \geq 0,$

$ch(indeterm_A) \geq 0,$

$ch(\bar{A}) \geq 0,$ for any $A \in NF$;

with the notations that "$indeterm_A$" means indeterminacy related to event $A$, and $\bar{A}$ is the opposite event of *A (*the *antiA* event*).*

*Second Axiom.*

The neutrosophic probability of the sample space is between ${}^-0$ and $3^+$.

$$NP(\nu\Omega) = \left(\sum_{x\in\nu\Omega} ch(x), ch(indeterm_{\nu\Omega}), ch(anti\ \nu\Omega)\right), \quad (56)$$

where

$$-0 \le \sum_{x\in\nu\Omega} ch(x) + ch(indeterm_{\nu\Omega}) + ch(anti\ \nu\Omega) \le 3^+, \quad (57)$$

with the notation $indeterm_{\nu\Omega}$ means total indeterminacy that may occur in the neutrosophic sample space.

For the classical complete (normalized) sample space,

*ch(anti νΩ)= 0,* but for incomplete sample space

$$ch(anti\ \nu\Omega) > 0. \quad (58)$$

*Third Axiom.*

This axiom is concerned with *neutrosophic σ-additivity:*

$$NP(A_1 \cup A_2 \cup \dots) =$$

$$= \left(\sum_{j=1}^{\infty} ch(A_j), ch\left(indeterm_{A_1 \cup A_2 \cup \ldots}\right), ch(\overline{A_1 \cup A_2 \cup \ldots})\right), \quad (59)$$

where $A_1, A_2, \ldots$ is a countable sequence of disjoint (or mutually exclusive) neutrosophic events.

If we relax the third axiom we get a *neutrosophic quasiprobability distribution*.

### 3.26. Consequences of Neutrosophic Probability Axioms.

*a) Monotonocity.*

If $A$ and $B$ are two neutrosophic events, with $A \subseteq B$, with

$$NP(A) = (ch(A), ch(indeterm_A), ch(\bar{A}))$$

$$NP(B) = \left(ch(B), ch(indeterm_B), ch(\bar{B})\right),$$

then

$$ch(A) \leq ch(B), \quad (60)$$

$$ch(indeterm_A) \leq ch(indeterm_B), \quad (61)$$

$$ch(\bar{A}) \geq ch(\bar{B}). \quad (62)$$

*b) Neutrosophic Probability of the Empty Set.*

$$NP(\emptyset) = (0,0,0). \tag{63}$$

*c) Bounding the Neutrosophic Probability.*

$$NP(A) = (ch(A), ch(indeterm_A), ch(\bar{A}))$$

where $0 \leq ch(A) \leq 1,$ (64)

$$0 \leq ch(indeterm_A) \leq 1, \tag{65}$$

$$0 \leq ch(\bar{A}) \leq 1. \tag{66}$$

*d) Neutrosophic Addition Law (or Neutrosophic Sum Rule):*

For any two neutrosophic events A and B we have:

$$NP(A \cup B) =$$

$$\big(ch(A) + ch(B) - ch(A \cap B), ch(indeterm_{A\cup B}), ch(\overline{A \cup B})\big). \tag{67}$$

If $A \cap B = \emptyset$, then

$$NP(A \cup B) = \big(ch(A) + ch(B), ch(indeterm_{A\cup B}), ch(\overline{A \cup B})\big). \tag{68}$$

*e) Neutrosophic Inclusion-Exclusion Principle.*

$$NP(\nu\Omega \setminus A) = \big(ch(\nu\Omega) - ch(A), ch\big(indeterm_{\nu\Omega\setminus A}\big), ch(A)\big). \tag{69}$$

Also, if $A \subseteq B$, then:

$$NP(B \setminus A) = \left(ch(B) - ch(A), ch\left(indeterm._{B\setminus A}\right), ch(\overline{B \setminus A})\right). \tag{70}$$

### 3.27. Interpretations of the Neutrosophic Probability.

Neutrosophic Probability can also have two interpretations, as the classical probability:

a) *Objective form*, or describing objective state of affairs, whose most popular version is the neutrosophic frequentist probability; and
b) *Subjective form*, or a degree of belief in an event to occur.

### 3.28. Neutrosophic Notions.

If an experiment produces indeterminacy, that is called a *neutrosophic experiment*. Collecting all results, including the indeterminacy, we get the *neutrosophic sample space* (or the *neutrosophic probability space*) of the experiment.

The *neutrosophic power set* of the neutrosophic sample space is formed by all different collections (that may or may not include the indeterminacy) of possible results. These collections are called *neutrosophic events*.

### 3.29. Example with Neutrosophic Frequentist Probability.

Let's consider a more concrete example.

Using the Frequentist Neutrosophic Probability we can (approximately of course) determine what is the chance that the die tosses as indeterminate. Similarly as in classical probability, we can use a computer simulation, based upon connections between *neutrosophic mathematical model* (i.e. models involving indeterminacy) and our everyday life. Neutrosophic statisticians can use simulations to approximate the probability of die uncertainty tossed on a specific irregular surface. With computers a large number of trials can be simulated in short time.

Suppose we obtain that the chance of getting indeterminacy $ch(indeterm)=0.10$ for tossing a regular die on an irregular surface. The neutrosophic sample space is then:

$$\nu\Omega=\{1,2,3,4,5,6,\ indeterm\}. \qquad (71)$$

Then, the neutrosophic probability of tossing event $A$ is

$$\mathcal{NP}(A)=\left(ch(A),ch(indeterm_A),ch(\overline{A})\right) \qquad (72)$$

where $ch(\cdot)$ mans "*ch*ance", and $\overline{A}$ is the opposite event of $A$ (chance that *antiA* occurs).

For example:

$$\begin{aligned}\mathcal{NP}(1)&=\left(ch(\{1\}),ch\left(indeterm_{\{1\}}\right),ch\left(\overline{\{1\}}\right)\right)\\&=\left(\frac{1-0.10}{6},0.10,5\cdot\frac{1-0.10}{6}\right)=(0.15,0.10,0.75)\\&=\mathcal{NP}(2)=\ldots=\mathcal{NP}(6).\end{aligned} \qquad (73)$$

In general:

$$\mathcal{NP}\left(\overline{A}\right)=\left(ch\left(\overline{A}\right),ch\left(indeterm_{\left(\overline{A}\right)}\right),ch\left(A\right)\right). \qquad (74)$$

Hence

$$\mathcal{NP}\left(\bar{1}\right)=\left(ch\left(\bar{1}\right),ch\left(indeterm_{\left(\bar{1}\right)}\right),ch\left(1\right)\right)$$

$$=\left(ch\left(\{2,3,4,5,6\}\right),ch\left(indeterm_{\left(\{2,3,4,5,6\}\right)}\right),ch\left(1\right)\right)$$

$$=\left(5\left(0.15\right),0.10,0.15\right)=0.75,0.10,0.15. \qquad (75)$$

Also, for

$$\mathcal{NP}\left(1\ \ or\ \ 2\right)=\left(ch\left(1\ \ or\ \ 2\right),ch\left(indeterm_{\left(1\ or\ 2\right)}\right),ch\left(\overline{1\ \ or\ \ 2}\right)\right)$$

$$=\left(ch\left(1\right)+ch\left(2\right),ch\left(indeterm_{\left(1\ or\ 2\right)}\right),ch\left(\overline{1\ \ and\ \ 2}\right)\right)$$

$$=\left(01.5+0.15,0.10,ch\left(\{3,4,5,6\}\right)\right)=\left(0.30,0.10,4\cdot\left(0.15\right)\right)$$
$$=\left(0.30,0.10,0.60\right). \qquad (76)$$

In general:

$$\mathcal{NP}\left(A\ \ or\ \ B\right)=\left(ch\left(A\right)+ch\left(B\right),ch\left(indeterm_{(A\ or\ B)}\right),ch\left(\overline{A\ \ or\ \ B}\right)\right) \qquad (77)$$

for $A\cap B=\phi$.

For *neutrosophic non-exclusive events* in general one has:

$$\mathcal{NP}\left(A \text{ or } B\right)=\left(ch\left(A \text{ or } B\right), ch\left(indeterm_{A \text{ or } B}\right), ch\left(\overline{A \text{ or } B}\right)\right)$$

$$= \left(ch(A)+ch(B)-ch(A \cap B), ch(indeterm), ch\left(\bar{A} and \bar{B}\right)\right).$$

(78)

Whence, if $A=\{1,2,3\}$, $B=\{2,3,4,5\}$, then:

$$\mathcal{NP}(\{1,2,3\} or \{2,3,4,5\}) =$$

$$= \left(3(0.15)+4(0.15)-2(0.15), 0.10, ch(\{4,5,6\} and \{1,6\})\right)$$

$$= \left(0.75, 0.10, ch(\{6\})\right) = (0.75, 0.10, 0.15).$$

In general, for independent events, one has:

$$\mathcal{NP}\left(A \text{ and } B\right)=\left(ch\left(A \text{ and } B\right), ch\left(indeterm_{A \text{ and } B}\right), ch\left(\overline{A \text{ and } B}\right)\right)$$

$$= \left(ch(A)\cdot ch(B), ch\left(indeterm_{A \text{ and } B}\right), ch\left(\overline{A \text{ and } B}\right)\right).$$

(79)

### 3.30. Example with Neutrosophic Frequentist Probability on a Neutrosophic Product Space.

Let suppose we toss the previous regular die on an irregular surface twice. Therefore we have two independent events. What is the neutrosophic probability of getting $\{3\}$ on the first tossing and *{4}* on the second tossing?

The first neutrosophic space with corresponding chances:

$$v\Omega_1 = \{\underset{0.15}{1,}\ \underset{0.15}{2,}\ \underset{0.15}{3,}\ \underset{0.15}{4,}\ \underset{0.15}{5,}\ \underset{0.15}{6,}\ \underset{0.10}{indeterm}\}, \qquad (80)$$

and the second neutrosophic space with corresponding chances:

$$v\Omega_2 = \{\underset{0.15}{1,}\ \underset{0.15}{2,}\ \underset{0.15}{3,}\ \underset{0.15}{4,}\ \underset{0.15}{5,}\ \underset{0.15}{6,}\ \underset{0.10}{indeterm}\} \qquad (81)$$

Whence we construct their *neutrosophic product space*:

$$\begin{Bmatrix} (1,1),(1,2),\ldots,(1,6) & (1,I),(2,I),\ldots,(6,I) & (I,I) \\ (1,1),(1,2),\ldots,(1,6) & (I,1),(I,2),\ldots,(I,6) & \end{Bmatrix}$$

$$\ldots\ldots\ldots\ldots\ldots\ldots\ldots\ldots\ldots\ldots \qquad (82)$$

$$(6,1),(6,2),\ldots,(6,6)$$

where $I = indeterminacy$,

with corresponding chances:

$$\begin{Bmatrix} 0.0225, 0.0225, \ldots, 0.0225 & 0.0150, 0.0150, \ldots, 0.0150 & 0.0100 \\ 0.0225, 0.0225, \ldots, 0.0225 & 0.0150, 0.0150, \ldots, 0.0150 & \end{Bmatrix}$$

$$\ldots\ldots\ldots\ldots\ldots\ldots\ldots\ldots\ldots\ldots$$

$$0.0225, 0.0225, \ldots, 0.0225$$

(83)

Hence,

$$ch(\{3\} and \{4\}) = 0.15(0.15) = 0.00225;$$

$$ch\left(indeterm_{\{3\}\ or\ \{4\}}\right) = 12(0.0150) + 0.0100 = 0.1800 + 0.0100$$
$$= 0.1900;$$

$$ch\left(\overline{\{3\} and \{4\}}\right) =$$
$$= \left(ch\left(\overline{\{3\}} \wedge \nu\Omega_2\right), \nu\Omega_1 \wedge \overline{\{4\}}, \{3\} \wedge \{1,2,3,5,6\}, \{1,2,4,5,6\} \wedge \{4\}\right)$$
$$= 35(0.0225) = 0.7875.$$

$$\mathcal{NP}(\{3\} and \{4\}) = (0.0225, 0.1900, 0.7875). \qquad (84)$$

We have considered that $(1,I),...,(6,I),(I,1),...,(I,6)$ are indeterminacies, while $(I,I)$ obviously is a double indeterminacy.

### 3.31. Example with Double Indeterminacy.

We change again the theoretical equipment. Instead of a fair die, we consider now a defect die in the sense that two of its faces have the print erased, for example the erased faces are $\{5\}$ and $\{6\}$.

The new neutrosophic probability space is:

$$\nu\Omega = \{1,2,3,4, indeterm_d, indeterm_s\} \qquad (85)$$

with two types of indeterminacies: one due to the physical die, denoted by $indeterm_d$, and the second due to the physical space, denoted by $indeterm_s$.

We consider that chance of $indeterm_s$ is the same as in the previous frequentist examples: $ch(indeterm_s) = 0.10$, and $ch(1) = ... = ch(4) = 0.15$ as before.

But from two erased prints we get $ch(indeterm_d) = 2(0.15) = 0.30$.

Thus

$$ch\left(total \quad indeterm\right) = ch(indeterm_s) + ch(indeterm_d)$$
$$= 0.10 + 0.30 = 0.40,$$

whence

$$\mathcal{NP}(1) = ... = \mathcal{NP}(4) = (0.15, 0.40, 0.45). \qquad (86)$$

This neutrosophic experiment is equivalent to experiment of having a perfect die with four faces (a tetrahedron), which is tossed on an irregular surface where the chance of indeterminacy (for the die to get stuck on one of its six edges or on one of its four vertices) is 0.40.

Therefore

$$v\Omega = \{1, 2, 3, 4, indeterm\}. \qquad (87)$$

### 3.32. Neutrosophic Example with Tossing a Coin Multiple Times.

Let's consider a regular coin [with two faces: H (head) and T (tale)] flipped on an irregular surface. By neutrosophic frequentist probability let's suppose the chance that the coin gets stuck on its edge into a surface crack is:

$$ch\ (indeterm) = 0.02 \quad (88)$$

Because the coin is fair, the chances of head or tale are equal:

$$ch\ (H) = ch(T) = \frac{1-0.02}{2} = 0.49. \quad (89)$$

The neutrosophic probability space is:

$$v\Omega = \{H, T, I\}, \quad (90)$$

where "*I*" stands for indeterm(inacy).

Therefore:

$$NP(H) = NP(T) = (0.49, 0.02, 0.49). \quad (91)$$

We flip the coin three times. What is the (neutrosophic) probability of getting *HTT*?

The neutrosophic product space is:

$$\begin{Bmatrix} H & T & I \\ 0.49 & 0.49 & 0.02 \end{Bmatrix} \times$$

$$\begin{Bmatrix} H & T & I \\ 0.49 & 0.49 & 0.02 \end{Bmatrix} \times$$

$$\begin{Bmatrix} H & T & I \\ 0.49 & 0.49 & 0.02 \end{Bmatrix}$$

$= \{\, HHH, HHT, HTH, HTT, THH, THT, TTH, TTT;$

and *indeterminacy of first order*:

$IHH, IHT, ITH, ITT;\ HIH, HIT, TIH, TIT; HHI, HTI, THI, TTI;$

also *indeterminacy of second order*:

$$IIH, IIT; IHI, ITI; HII, TII;$$

and *indeterminacy of third order*:

$III\ \}$, (92)

which has $3^3 = 27$ elements.

Computing the chances:

$$ch(HHH) = ch(HHT) = \cdots = ch(TTT) = (0.49)^3 = 0.117649;$$

$$ch(IHH) = ch(IHT) = \cdots = ch(TTI) = (0.49)^2(0.02) = 0.004802,$$

for each first order indeterminacy;

$$ch(IIH) = ch(IIT) = \cdots = ch(TII) = 0.49(0.02)^2 = 0.000196,$$

for each second order indeterminacy;

$$ch(III) = (0.02)^3 = 0.000008.$$

Therefore indeterminacy propagates.

The sum of total indeterminate chances is:

$$ch(total\ indeterm) = 12(0.004802) + 6(0.000196) + 1(0.000008) = 0.058808.$$

The chance that *HTT* occurs is

$$ch(HTT) = (0.49)^3 = 0.117649,$$

while the chance that HTT does not occur is:

$$ch(\overline{HTT}) = 7(0.117649) = 0.823543.$$

Finally,

$$NP(HTT) = (0.117649, 0.058808, 0.823543).$$

In the classical probability, where $ch(indeterm) = 0$,

we get $P(HTT) = 0.5^3 = 0.125$,

and, transcribed into neutrosophic form, we get:

$$NP(HTT) = (0.5^3, 0, 7(0.5)^3) = (0.125, 0, 0.875). \quad (93)$$

The chance of flipping three times in a row and getting *HTT* is smaller in the neutrosophic probability space than in the classical probability space, because of the strictly positive chance of having indeterminacy:

$$0.117649 < 0.125000. \quad (94)$$

## 3.33. Example with Sum of Chances of an Event.

In *classical probability:* If $A$ is an event, then $P(A)$ is the sum of probabilities of all outcomes in the set $A$.

In *neutrosophic probability*, it is similar if

$$A = \{a_1, a_2, \dots, a_n\}. \quad (95)$$

$NP(A) =$ **(** sum of chances of all outcomes in the set $A, ch(indeterm_A), ch(\overline{A})$ **)** =

$$\left(\sum_{j=1}^{n} ch(a_j), ch(indeterm_A), ch(\bar{A})\right). \quad (96)$$

For example, if we retake one of the previous experiments of a regular die tossed on an irregular surface, where the chance of indeterminacy is *0.10*, then

$$NP(\{1,2,3\}) = \left(ch\{1,2,3\}, ch\left(indeterm_{\{1,2,3\}}\right), ch(\overline{\{1,2,3\}})\right)$$

$$= \left(ch(1) + ch(2) + ch(3), ch\left(indeterm_{\{1,2,3\}}\right), ch(\{4,5,6\})\right)$$

$$= (0.15 + 0.15 + 0.15, 0.10, ch(4) + ch(5) + ch(6)$$

$$= (0.45, 0.10, 0.45), \quad (97)$$

since $NP(1) = NP(2) = NP(3) = (0.15, 0.10, 0.75)$.

## 3.34. Paraconsistent Neutrosophic Probability.

The paraconsistent neutrosophic probability has the property that the sum of its components is strictly greater than *1*:

$$t + i + f \in (1, 3^{+}[, \tag{98}$$

therefore one has contradictions between chances.

Forecasting an event from different criteria, we may obtain different chances of occurrence.

For example, suppose two handball teams *G* and *H* will compete in a game next week.

a) According to the *history* of their previous disputes, team *G* is *60%* favorable to win.
b) But, according to their *last games in the actual season vs. other handball teams*, *H* is showing a better performance than *G*, and the experts conclude upon this criterion that *H* has *70%* chance to win.
c) Others believe that since *G* was often better than *H*, but in this season *H* contrarily played better than *G*, as a compensation it is *10%* chance that their game be undecided (tie).

Therefore, *NP(G wins over H) = (0.6, 0.1, 0.7),*

with *0.6 + 0.1 + 0.7 > 1.* *(99)*

## 3.35. Incomplete Neutrosophic Probability.

The incomplete neutrosophic probability has the property that the sum of its components is strictly less than *1*:

$$t + i + f \in ]^{-}0,1), \qquad (100)$$

therefore one has incomplete (missing) information.

Lets' reconsider the previous example about two handball teams *H* and *G* that will compete in a game next week.

a) If both teams have a weak performance in the present season and of almost equal values, then each one will have a slim chance to win on *20%.*
b) Studying the low number of their previous games when the results were tie, the handball experts conclude that it is a slim chance of *30%* of having a tie game.

Therefore, *NP(G wins over H) = (0.2, 0.3, 0.2),*

with *0.2 + 0.3 + 0.2 < 1.* *(101)*

## 3.36. Neutrosophic Mutually Exclusive Events.

In *classical probability*, if $A$ and $B$ are *mutually exclusive (independent) events*, then

$$P(A\ or\ B) = P(A) + P(B). \tag{102}$$

In *neutrosophic probability*, we have similar property for mutually exclusive events:

$$NP(A\ or\ B) = \big(ch(A) + ch(B), ch(indeterm_{A\ or\ B}), ch\ (\overline{A\ or\ B})\big). \tag{103}$$

In *classical probability* for *non mutually exclusive events* $A$ and $B$ one has:

$$P(A \text{ or } B) = P(A) + P(B) - P(A \text{ and } B). \tag{104}$$

In *neutrosophic probability* for *non mutually exclusive neutrosophic events* one similarly has:

$$NP(A \text{ or } B) = \big(ch(A) + ch(B) - ch(A \text{ and } B), ch(indeterm_{A\ or\ B}), ch(\overline{A \text{ or } B})\big). \tag{105}$$

For example, let's consider a deck of *52* cards, but such that *2* of them are deteriorated and one cannot read them. Let's draw at random a single card. What is the neutrosophic probability of getting a face card (event *A*) or a heart card (event *B*)? We know that none of the face and heart cards were deteriorated. There are *12* face cards (four types of each of *J, Q,* and *K*), *13* heart cards, and *3* cards that are both face and heart.

$$NP(A\ or\ B) =$$
$$(ch(A\ or\ B), ch(indeterm_{A\ or\ B}), ch(\overline{A\ or\ B})) =$$
$$(ch(A) + ch(B) -$$
$$ch\ (A\ and\ B), ch(indeterm_{A\ or\ B}), ch\ (\bar{A}\ and\ \bar{B})) =$$
$$\left(\frac{12}{52} + \frac{13}{52} - \frac{3}{52}, \frac{2}{52}, \frac{52-12-13+3-2}{52}\right) = \left(\frac{22}{52}, \frac{2}{52}, \frac{28}{52}\right). \quad (106)$$

Of course,

$$NP(A) = \left(\frac{12}{52}, \frac{2}{52}, \frac{38}{52}\right),$$

$$NP(B) = \left(\frac{13}{52}, \frac{2}{52}, \frac{37}{52}\right),$$

$$NP(A\ and\ B) = \left(\frac{3}{52}, \frac{2}{52}, \frac{47}{52}\right). \quad (107)$$

We do not simplify the fractions because we can better compare these neutrosophic probabilities if we leave the same denominators for them all.

But let's say we don't know if any of the two erased cards are among the face or heart cards. Then:

$$NP(A) = \left(\left[\frac{10}{52}, \frac{12}{52}\right], \frac{2}{52}, \left[\frac{38}{52}, \frac{40}{52}\right]\right), \quad (108)$$

and $NP(B) = \left(\left[\frac{11}{52}, \frac{13}{52}\right], \frac{2}{52}, \left[\frac{37}{52}, \frac{39}{52}\right]\right)$, (109)

and $NP(A \text{ and } B) = \left(\left[\frac{1}{52}, \frac{3}{52}\right], \frac{2}{52}, \left[\frac{47}{52}, \frac{49}{52}\right]\right)$, (110)

whence $NP(A \text{ or } B) = \left(\left[\frac{18}{52}, \frac{24}{52}\right], \frac{2}{52}, \left[\frac{26}{52}, \frac{32}{52}\right]\right)$, (111)

because $ch(A \text{ or } B) =$

$$= \left[\frac{10}{52}, \frac{12}{52}\right] + \left[\frac{11}{52}, \frac{13}{52}\right] - \left[\frac{1}{52}, \frac{3}{52}\right]$$
$$= \left[\frac{21}{52}, \frac{25}{52}\right] - \left[\frac{1}{52}, \frac{3}{52}\right]$$
$$= \left[\frac{21-3}{52}, \frac{25-1}{52}\right] = \left[\frac{18}{52}, \frac{24}{52}\right],$$

and $ch(\overline{A \text{ or } B}) =$

$$= ch(whole\ neutrosophic\ probability\ space) - ch(indeterm) - ch(A\ or\ B)$$

$$= 1 - \frac{2}{52} - \left[\frac{16}{52}, \frac{24}{52}\right] = \frac{50}{52} - \left[\frac{18}{52}, \frac{24}{52}\right] = \left[\frac{26}{52}, \frac{32}{52}\right].$$

## 3.37. Neutrosophic Experimental Probability.

In *classical experimental probability* is

$$\frac{number\ of\ times\ event\ A\ occurs}{total\ number\ of\ trials}. \quad (112)$$

Similarly, *Neutrosophic Experimental Probability* is:

$$\left(\begin{matrix}\frac{number\ of\ times\ event\ A\ occurs}{total\ number\ of\ trials}, \quad \frac{number\ of\ times\ indeterminacy\ occurs}{total\ number\ of\ trials}, \\ \frac{number\ of\ times\ event\ A\ does\ not\ occur}{total\ number\ of\ trials}\end{matrix}\right).$$

(113)

### 3.38. Neutrosophic Survey.

A *Neutrosophic Survey* is a way to obtain neutrosophic experimental probability.

Example. Let's say that we toss a regular die five times on an irregular surface, and we get: $2, 5, 1, indeterminacy, 4$.

### 3.39. Neutrosophic Conditional Probability for Independent Events.

In *classical probability*, if $A$ and $B$ are *independent events*, then $P(A\ given\ B) = P(A)$. (114)

Similarly for *neutrosophic independent events*:

$NP(A\ given\ B) = NP(A),$ (115)

because

$$ch(A\ given\ B) = ch(A),$$
$$ch(indeterm_A\ given\ B) = ch(indeterm_A), \qquad (116)$$

and

$$ch\ (\bar{A}\ given\ B) = ch(\bar{A}). \qquad (117)$$

## 3.40. Neutrosophic Probability of an Impossible Event (Φ) on the neutrosophic probability space $\nu\Omega$ is:

$$NP(\Phi) = \begin{pmatrix} 0,\ ch(indeterm), \\ ch(\nu\Omega) - ch(indeterm) \end{pmatrix}. \qquad (118)$$

## 3.41. Neutrosophic Probability of a Sure Event ($\nu\Omega$) on the neutrosophic probability space $\nu\Omega$ is:

$$NP(\nu\Omega) = (1 - ch(indeterm), ch(indeterm), 0). \qquad (119)$$

## 3.42. Neutrosophic Bayesian Rule.

In *classical probability*, the Bayesian Rule is:

$$P(A|B) = P(B|A)\frac{P(A)}{P(B)}. \tag{120}$$

Let's examine the neutrosophic version of this rule.

Suppose we have an urn with *5 A*-votes, *2* indeterminate (unclear, erased) votes, and *3 B*-votes.

If $A$ is the event of extracting an $A$-vote from the urn, and $B$ the event of extracting a $B$-vote from the urn, then:

$$NP(A) = \left(\frac{5}{10}, \frac{2}{10}, \frac{3}{10}\right), NP(B) = \left(\frac{3}{10}, \frac{2}{10}, \frac{5}{10}\right). \tag{121}$$

If one B-vote has be taken from the urn, then

$$NP(A|B) = \left(\frac{5}{9}, \frac{2}{9}, \frac{2}{9}\right). \tag{122}$$

But if one $A$-votes has been taken from the urn, then

$$NP(B|A) = \left(\frac{3}{9}, \frac{2}{9}, \frac{4}{9}\right). \tag{123}$$

In general, the *Neutrosophic Bayesian Rule* is:

$$NP(A|B) = \left(ch(A|B), ch(indeterm_A \mid B), ch(\bar{A}|B)\right)$$

$$= \left( ch(B|A)\frac{ch(A)}{ch(B)}, ch(indeterm_A \mid B), ch\ (\bar{A}|B) \right). \quad (124)$$

Therefore, as in classical probability

$$ch(A|B) = ch(B|A)\frac{ch(A)}{ch(B)}. \quad (125)$$

For our particular example, we get:

$$NP(A|B) =$$

$$= \left( ch(B|A)\frac{ch(A)}{ch(B)}, ch(indeterm_A|B), ch(\bar{A}|B) \right)$$

$$= \left( \frac{3}{9} \times \frac{\frac{5}{10}}{\frac{3}{10}}, \quad \frac{2}{9}, \quad ch(B|B) \right)$$

$$= \left( \frac{5}{9}, \frac{2}{9}, \frac{2}{9} \right). \quad (126)$$

## 3.43. Neutrosophic Multiplicative Rule.

In *classical probability*, the *Multiplication Rule for Probabilities* (equivalent with the *Conditional Probability*) is:

$$P(A\ and\ B) = P(A) \cdot P(B\ given\ A). \quad (127)$$

The *Multiplication Rule for Neutrosophic Probabilities* is:

$$NP(A\ and\ B) = (\ ch(A) \cdot ch(B\ given\ A),$$
$$ch(indeterm_{A\ and\ B}) + ch(indeterm_{A\ or\ B}\ |\ A) -$$
$$ch(indeterm_{A\ and\ B}) \cdot ch(indeterm_{A\ and\ B}|A),$$
$$ch(A) \cdot ch(A\ given\ A) + ch(B) \cdot ch(A\ given\ B) +$$
$$ch(B) \cdot ch(B\ given\ B)\big), \tag{128}$$

because:

$$\begin{aligned}
ch\big(indeterm\ for\ (A\ and\ B)\big) = \\
&= ch(indeterm) \cdot ch(indeterm|A) \\
&+ ch(indeterm) \cdot ch(A|A) \\
&+ ch(indeterm) \cdot ch(B|A) \\
&+ ch(indeterm|A) \cdot ch(A) \\
&+ ch(indeterm|A) \cdot ch(B) \\
&= ch(indeterm|A) \\
&\cdot \Big[\underbrace{ch(indeterm) + ch(A) + ch(B)}\Big] \\
&+ ch(indeterm) \\
&\cdot \Big[\underbrace{ch(A|A) + ch(B|A) + ch(indeterm|A)} \\
&- ch(indeterm|A)\Big] \\
&= ch(indeterm) + ch(indeterm|A) \\
&- ch(indeterm) \cdot ch(indeterm|A),
\end{aligned}$$

due to the facts that

$$\left[\underbrace{ch(indeterm) + ch(A) + ch(B)}\right] = 1 \qquad (129)$$

and $\left[\underbrace{ch(A|A) + ch(B|A) + ch(indeterm|A)}\right] = 1.$

(130)

Let's consider the previous neutrosophic example:

$$\underset{A - votes}{5} \quad \underset{indeterm - votes}{2} \quad \underset{B - votes}{3}$$

We pick two votes in succesion without replacement.

Suppose $A$ is the event that the first is an A-vote, and B is a B-vote.

We have:

$$ch(A) = \frac{5}{10}, ch(indeterm) = \frac{2}{10},$$

$$ch(B) = \frac{3}{10}, ch(A|A) = \frac{4}{9},$$

$$ch(indeterm|A) = \frac{2}{9}, ch(B|A) = \frac{3}{9},$$

$$ch(A|B) = \frac{5}{4}, ch(B|B) = \frac{2}{9},$$

whence:

$NP(A \text{ and } B) = \left(\frac{5}{10}\cdot\frac{3}{9}, \frac{2}{10}+\frac{2}{9}-\frac{2}{10}\cdot\frac{2}{9}, \frac{5}{10}\cdot\frac{4}{9}+\frac{3}{10}\cdot\frac{5}{9}+\frac{3}{10}\cdot\frac{2}{9}\right) = \left(\frac{15}{90}, \frac{34}{90}, \frac{41}{90}\right).$ (131)

## 3.44. Neutrosophic Negation (or Neutrosophic Probability of Complementary Events).

For any event *A* different from indeterminacy, from the sample space *X*, one has:

$NP(A) = ( ch(A), ch(indeterm_A), ch(antiA) ),$ (132)

whence the neutrosophic probability of the complement of *A*, noted as $\mathcal{C}(A)$ (or as *antiA*) is:

$NP( \mathcal{C}(A) ) = NP( antiA ) = ( ch(antiA), ch(indeterm_{antiA}),$

$ch( anti(antiA) ) = ( ch(X)-ch(A), ch(indeterm_{antiA}), ch(A) ).$ (133)

## 3.45. De Morgan's Neutrosophic Laws.

$NP(\mathcal{C}(A \cup B) ) = ( ch(\mathcal{C}(A \cup B) ), ch(indeterm_{\mathcal{C}(A \cup B)}),$

$ch( \mathcal{C}(\mathcal{C}(A \cup B)) )$

$= ( ch(\mathcal{C}(A)\cap\mathcal{C}(B) ), ch(indeterm_{\mathcal{C}(A)\cap\mathcal{C}(B)}),$

$ch( \mathcal{C}(\mathcal{C}(A)\cap\mathcal{C}(B)) )$

$= NP(\mathcal{C}(A)\cap\mathcal{C}(B)).$ (134)

Similarly,

$$NP(\mathcal{C}(A\cap B)) = (\, ch(\mathcal{C}(A\cap B)\,), ch(indeterm_{\mathcal{C}(A\cap B)}),$$

$$ch(\,\mathcal{C}(\mathcal{C}(A\cap B))\,)$$

$$= (\, ch(\mathcal{C}(A) \cup \mathcal{C}(B)\,), ch(indeterm_{\mathcal{C}(A) \cup \mathcal{C}(B)}),$$

$$ch(\,\mathcal{C}(\mathcal{C}(A) \cup \mathcal{C}(B))\,)$$

$$= NP(\mathcal{C}(A) \cup \mathcal{C}(B)). \qquad (135)$$

## 3.46. Neutrosophic Double Negation.

In classical probability,

$$P(\,anti(antiA)\,) = P(A). \qquad (136)$$

In neutrosophic probability, for $A$ an event different from indeterminacy:

$$NP(A) = (\, ch(A), ch(indeterm_A), ch(antiA)\,), \qquad (137)$$

then:

$$NP(antiA) = (\, ch(antiA), ch(indeterm_{antiA}), ch(anti(antiA))\,)$$

$$= (\, ch(antiA), ch(indeterm_{antiA}), ch(A)\,) \qquad (138)$$

whence

$$NP(\,anti(antiA)\,) = (\, ch(anti(antiA)), ch(indeterm_{anti(antiA)}), ch(antiA)\,) = (\, ch(A), ch(indeterm_A), ch(antiA)\,) = NP(A). \quad (139)$$

Let's reconsider the previous example about a urn with:

$$\underset{A-votes}{5} \quad \underset{indeterm-votes}{2} \quad \underset{B-votes}{3}$$

*NP(A)=(5/10, 2/10, 3/10),*

then *NP( antiA ) = (3/10, 2/10, 5/10),*

and it follows that *NP( anti(antiA) )=(5/10, 2/10, 3/10) = NP(A).*

### 3.47. Neutrosophic Expected Value.

Let's consider a neutrosophic discrete probability space $X$ with the determined outcomes $x_1, x_2, \ldots, x_r$ and their respective chances to occur $p_1, p_2, \ldots, p_r,$ and with indeterminacies $indeterm_1, indeterm_2, \ldots, indeterm_k$, then the Neutrosophic Expected Value (NE) is:

$$NE = \sum_{j=1}^{r} n_j p_j + \sum_{k=1}^{s} (m_k \cdot \mathrm{ch}(indeterm_k)) \quad (140)$$

where $n_j$ is the possible numerical outcome for the corresponding chance $p_j$, for all $j$, and $m_k$ is the possible numerical outcome for the corresponding chance that $indeterm_k$ occurs, for all $k$.

If we reconsider the previous neutrosophic example:

$$\underset{A - votes}{5} \quad \underset{indeterm - votes}{2} \quad \underset{B - votes}{3}$$

And the numerical outcomes for extracting an $A$-vote is loosing \$2.00, for extracting a $B$-vote is gaining *\$3.00*, while for extracting an indeterminate vote is loosing \$1.00. What is the neutrosophic expected value?

$$NE = -2\times(5/10) + 3\times(3/10) - 1\times(2/10) = -\$0.30. \quad (141)$$

## 3.48. Neutrosophic Probability and Neutrosophic Logic Used in The Soccer Games.

For all games where there are three possible results (winning, loosing, or tie), neutrosophic probability works perfectly, but the classical or imprecise probabilities do not apply, since they can describe one result only.

Let's say: What is the probability that a soccer team wins in a soccer game? Neutrosophic probability gives all three chances: chance to win, chance to get a tie game, and chance to loose.

Suppose two soccer games will play: teams *Alpha* (α) vs. *Beta* (β), and *Gamma* (γ) vs. *Delta* (δ).

$$NP(Alpha\ to\ win) = (0.7, 0.2, 0.1), \quad (142)$$

which means that *Alpha* has 0.6 chance to win, 0.2 chance to tie game, and 0.1 chance to loose,

$$NP(Gamma\ to\ win) = (0.3, 0.5, 0.2), \qquad (143)$$

then what is the neutrosophic probability that both teams *Alpha* and *Gamma* win in their soccer games?

We make the product of the neutrosophic probability spaces:

$$\begin{Bmatrix} W_\alpha, & I_{\alpha\beta}, & L_\alpha \\ 0.7 & 0.2 & 0.1 \end{Bmatrix} \times \begin{Bmatrix} W_\gamma, & I_{\gamma\delta}, & L_\gamma \\ 0.3 & 0.5 & 0.2 \end{Bmatrix} \qquad (144)$$

where $W_\alpha = \alpha$ winning, $I_{\alpha\beta} =$ indeterminacy (tie games between $\alpha$ and $\beta$), $L_\alpha =$ $\alpha$ loosing; similarly for $W_\gamma, I_{\gamma\delta}, L_\gamma$, which is

$$\begin{Bmatrix} W_\alpha W_\gamma, & W_\alpha I_{\gamma\delta}, & W_\alpha L_\gamma, & I_{\alpha\beta} W_\gamma, & I_{\alpha\beta} I_{\gamma\delta}, & I_{\alpha\beta} L_\gamma, & L_\alpha W_\gamma, & L_\alpha I_{\gamma\delta}, & L_\alpha L_\gamma \\ 0.21 & 0.35 & 0.14 & 0.06 & 0.10 & 0.04 & 0.03 & 0.05 & 0.02 \end{Bmatrix}, \qquad (145)$$

and the numbers below each possible outcome represent their corresponding chances to occur.

We can re-arrange the final result in many ways.

a) In the classical probability, we can say:

$$P\big((Alpha\ winning)\&(Gamma\ winning)\big) = 0.7(0.3) = 0.21, \qquad (146)$$

while $1 - 0.21 = 0.79$ is the probability of the opposite event, negation of $\big((Alpha\ winning)\&(Gamma\ winning)\big)$, i.e. in the two soccer games either there is at least a tie game, or at least one of the teams *Alpha* or *Gamma* looses.

b) In the neutrosophic probability, <u>the outcome is more refined</u>.

$$b1)\quad NP\big((Alpha\ winning)\&(Gamma\ winning)\big) = \{\, ch(Alpha\ Winning\ \&\ Gamma\ winning),$$

$$ch(at\ least\ one\ tie\ game),$$

$$ch(\text{referring to } Alpha \text{ and } Gamma\text{: one loosing, the other winning, or both loosing})\} =$$

$$= (0.21, 0.35 + 0.06 + 0.10 + 0.04 + 0.05, 0.14 + 0.03 + 0.02) = (0.21, 0.60, 0.19). \qquad (147)$$

$$b2)\quad NP\big((Alpha\ winning)\&(Gamma\ winning)\big) =$$

$$\{\, (Alpha\ winning)\&(Gamma\ winning),$$

$$ch(at\ least\ one\ of\ Alpha\ and\ Gamma\ has\ a\ tie\ game, and\ none\ looses),$$

$$ch(at\ least\ one\ of\ Alpha\ or\ Gamma\ looses)) = \\ = (0.21, 0.35 + 0.06 + 0.10, 0.14 + 0.04 + 0.03 + 0.05 + 0.02) = (0.21, 0.51, 0.28). \quad (148)$$

c) Another solution to this soccer game would be to use neutrosophic logic. Let's consider

$P_1 = \{\text{Team } Alpha \text{ will win}\}$,

and $P_2 = \{\text{Team } Gamma \text{ will win}\}$

as *two neutrosophic logical propositions* whose values are (0,7, 0.2, 0.1), and respectively (0.3, 0.5, 0.2).

Then one uses the neutrosophic operator "and" $(\Lambda_N)$ as part of the $\text{N} - \text{norm}$:

$$P_1 \Lambda_N P_2 = (0.7 \Lambda_F 0.3, 0.2 \text{V}_F 0.5, 0.1 \text{V}_F 0.2), \quad (149)$$

where $\Lambda_F$ is the fuzzy "and" operator (t-norm),

and $\text{V}_F$ is the fuzzy "or" operator (t-conorm).

c1. If we take the fuzzy and/or operators min/max, we get:

$$P_1 \Lambda_N P_2 = (\min(0.7, 0.3), \max(0.2, 0.5), \max(01., 0.2)) = (0.3, 0.5, 0.2). \quad (150)$$

c2. If we take the fuzzy *and/or* operators as $x \cdot y \,/\, x + y - x \cdot y,$ (151)

we get

$$P_1 \Lambda_N P_2 = \big(0.7(0.3), 0.2 + 0.5 - 0.2(0.5), 0.1 + 0.2 - 0.1(0.2)\big) = (0.21, 0.60, 0.28). \quad (152)$$

(In neutrosophic logic, the sum of its three components may be different from 1.)

Similarly for other particular t-norms/t-conorms.

### 3.49. A Neutrosophic Question.

Rolling two regular dice on an irregular surface, what is the neutrosophic probability of getting a sum of 6?

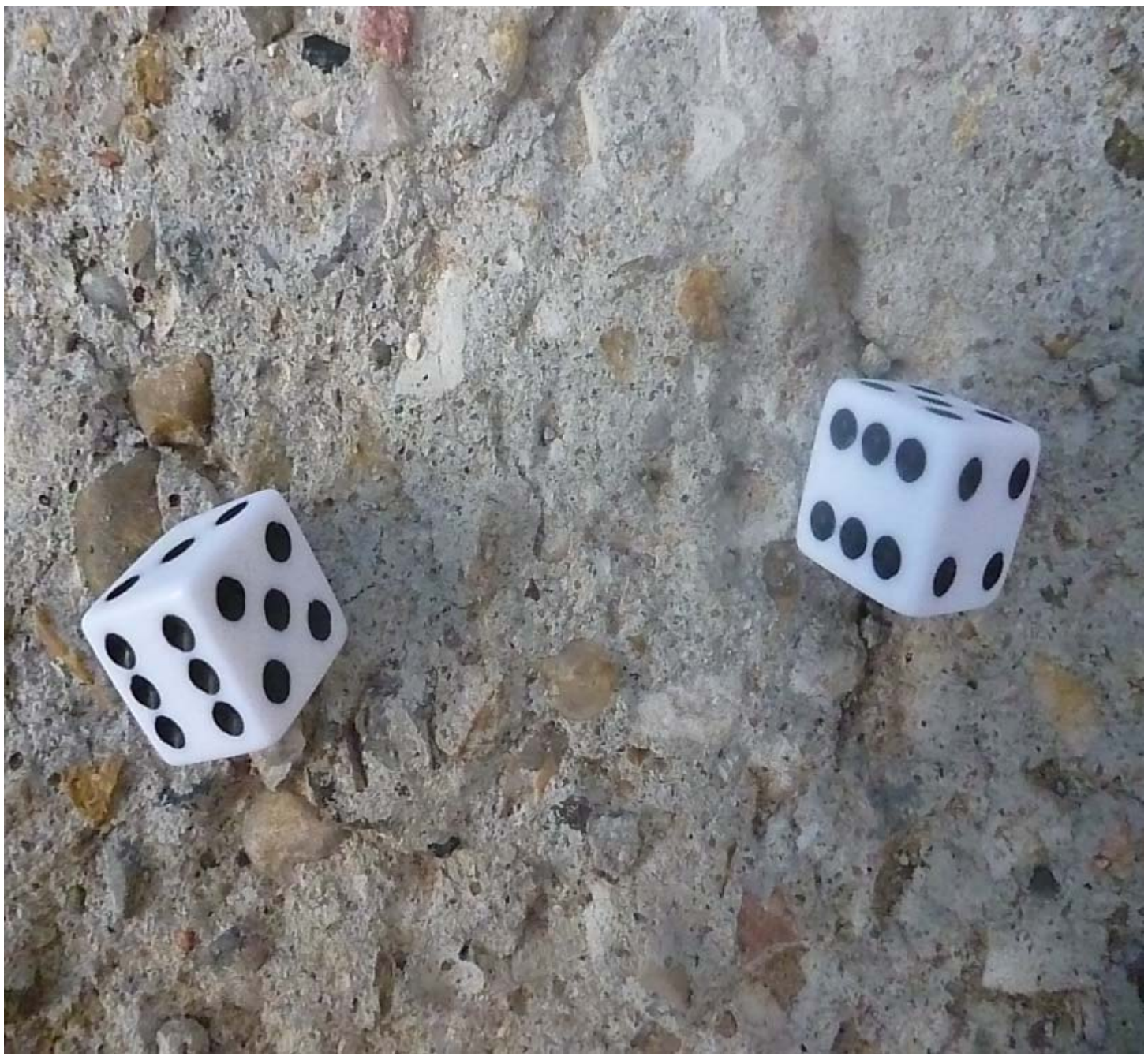

*Fig. 9. Double Indeterminacy*

The five favorable cases will be:

$1 + 5, 2 + 4, 3 + 3, 4 + 2, 5 + 1.$ (153)

But what about:

$$6 + indeterm, \text{ and } indeterm + 6?$$

Should we consider that

$$6 + indeterm = 6 \qquad (154)$$

(and consequently: $indeterm + 6 = 6$) ?

or should we say that

$$6 + indeterm = indeterm \ ?$$

Of course,

$$indeterm + indeterm = indeterm. \qquad (155)$$

Surely, in a game the players can make conventions among themselves, for example that a number plus indeterminacy is equal to that number, but this would mean that indeterminacy is taken for zero, which is not quite true.

Let's compute $NP(sum = 6)$.

Neutrosophic Probability Spaces are:

$\nu\Omega_1 = \{1, 2, 3, 4, 5, 6, indeterm_1\}$ for die #1;

$\nu\Omega_2 = \{1, 2, 3, 4, 5, 6, \text{indeter}m_2\}$ for die # 2.

The neutrosophic probability product space is

$$\Omega_1 \times \nu\Omega_2$$
$$= \{ (1,1), (1,2), \dots, (1,6), \dots (6,1), (6,2), \dots, (6,6),$$

$$(\text{indeter}m_1, 1), (\text{indeter}m_1, 2), \dots, (\text{indeter}m_1, 6),$$

$$(1, \text{indeter}m_2), (2, \text{indeter}m_2), \dots, (6, \text{indeter}m_2),$$

$(\text{indeter}m_1, \text{indeter}m_2)$ }. (156)

Considering that through frequentist neutrosophic experiment for a single die we have found the chance of getting indeterminacy is 0.10, whence $ch(1) = \cdots = ch(6) = \frac{1-0.1}{6} = 0.15,$ we get:

$NP(\text{sum} = 6) == \big(ch(\text{sum} = 6), ch(\text{indeter}m_{sum=6}), ch(\text{sum} \neq 6 \text{ and no indeterm})\big) = (\,5 \cdot (0.15)(0.15), 12(0.10)(0.15) + 0.10(0.10), 31(0.15)(0.15)\,) = (0.1125, 0.1900, 0.6975).$ (157)

In classical probability, where there is no indeterminacy,

$$P(\text{sum} = 6) = \frac{5}{36} \approx 0.1389 > 0.1125$$

$= ch(sum = 6)$ (158)

from neutrosophic probability.

## 3.50. Neutrosophic Discrete Probability Spaces.

In general, if we have two neutrosophic discrete probability spaces, with the chances of occuring of each event listed below that event

$$\nu\Omega_1 = \begin{matrix} \{A_1, & A_2, & \dots, A_{\mathrm{n}}, & \mathrm{indeterm}_1\} \\ p_1, & p_2, \dots, & p_{\mathrm{n}}, & \mathrm{ch}(\mathrm{indeterm}_1) \end{matrix}$$

and

$$\nu\Omega_2 = \begin{matrix} \{B_1, & B_2, & \dots, B_{\mathrm{n}}, & \mathrm{indeterm}_2\} \\ q_1, & q_2, \dots, & q_{\mathrm{n}}, & \mathrm{ch}(\mathrm{indeterm}_2), \end{matrix}$$

then the neutrosophic probability of having event $A_j$ and event $B_k$ to occur is:

$$NP(A_j \text{ and } B_k) = \left(p_j \cdot q_k, I_1 + I_1 - I, I_2, \sum_{\substack{u=1,n \\ v=1,n}} p_u q_v - p_j \cdot q_k\right). \quad (159)$$

This can be further generalized to the neutrosophic discrete probability product of $s$ spaces:

$$\nu\Omega_1 \times \nu\Omega_2 \times \dots \times \nu\Omega_{\mathrm{s}} = \prod_{\mathrm{r}=1}^{\mathrm{s}} \{A_{\mathrm{r},1}, A_{\mathrm{r},2}, \dots, \mathrm{A}_{\mathrm{r},\mathrm{n}_\mathrm{r}}, \mathrm{indeterm}_{\mathrm{r}}\}$$

with corresponding neutrosophic probabilities

$P_{r,1} \quad P_{r,2} \quad \dots \quad P_{r,n_r} \quad I_r.$

Then $NP\left(A_{1,j_1} \text{ and } A_{2,j_2} \text{ and} \ldots \text{ and } A_{s,j_s}\right) =$

$$\left( \prod_{r=1}^{s} p_{r,j_r}, \sum_{t=1}^{s} (-1)^{t-1} S_t, \sum_{\substack{k_1=1,n_1 \\ k_2=1,n_2 \\ \cdots\cdots \\ k_s=1,n_s}} \left( \prod_{r=1}^{s} p_{r,k_r} \right) - \prod_{r=1}^{s} p_{r,j_r} \right), \qquad (160)$$

where

$$S_1 = I_1 + I_2 + \cdots + I_s \qquad (s \text{ terms} = C_s^1)$$

$$S_2 = I_1 I_2 + I_1 I_3 + \cdots + I_1 I_s + \cdots + I_{s-1} I_s \qquad \left( \frac{s(s-1)}{2} \text{ terms} = C_s^2 \right)$$

………………………………………………………

$$S_t = \sum_{(j_1, j_2, \ldots, j_t) \in \mathcal{C}_{\{1,2,\ldots,s\}}^t} I_{j_1} I_{j_2} \ldots I_{j_t} \qquad \left( \frac{s!}{t!(s-t)!} \text{ terms} = C_s^t \right) \qquad (161)$$

where $\mathcal{C}_{\{1,2,\ldots,s\}}^t$ is the family of all subsets of $\{1, 2, \ldots, s\}$, such that the cardinal of each subset is $t$, for $1 \le t \le s, t$ integer.

## 3.51. Classification of Neutrosophic Probabilities.

1. There is an *objective neutrosophic probability* when the chances of all events, including the chance of identerminacy, can be computed objectively.

For example: Tossing a cubic die, which has two faces that are unreadable, on a regular surface. Let's consider that the numbers *5* and *6* are erased. We can exactly compute

$$ch(\text{indeterm}) = \frac{2}{6},$$

$$ch(1) = ch(2) = ch(3) = ch(4) = \frac{1}{6}.$$

2. The *frequentist neutrosophic probability* when at least the chance of one event, or the chance of some indeterminacy, cannot be computed objectively (exactly), but one can make experiments in order to compute frequentist chances.

For example: Tossing a regular die on an irregular surface having many cracks. We are not able to exactly compute the chance of indeterminacy (i.e when the die gets stuck in a crack on a vertex or on an edge). We may experiment, tossing the die a number of times in order to compute chance of indeterminacy as number of favorable cases over total number of cases. But this is just an aproximation. And if we repeat the experiment, we get a different result.

3. *Subjective Neutrosophic Probability* is neither posssible to compute it objectively (exactly), nor to experiment it and compute it as frequentist chance.

For example, and aircraft is detected in the sky. A source estimates it as to be 60% friend, 30% hostile, and 10% neutral. The estimation is subjective. Another source could give us a different estimation.

## 3.52. The Fundamental Neutrosophic Counting Principle.

Let's consider a neutrosophic event $E$ that can occur in $e$ ways and $e_I$ indeterminacies. After $E$ has occurred, a neutrosophic event $F$ can occur in $f$ ways and $f_I$ indeterminacies. Then, the neutrosophic event $E$ followed by the neutrosophic event $F$ can occur in $e \cdot f$ ways, and in $e_I \cdot f + e \cdot f_I$ indeterminacies of first order, and in $e_I \cdot f_I$ indeterminacies of second order.

Taking the previous example about tossing a cubic die, which has two faces *5* and *6* that are unreadable, but on an irregular surface. And then another cubic die with all readable faces, tossed on an irregular surface. We have the following neutrosophic sample spaces:

$$_N\Omega_1 = \{1, 2, 3, 4, indeterm_{die}, indeterm_{space}\} \quad (162)$$

$$_N\Omega_2 = \{1, 2, 3, 4, 5, 6, indeterm_{space}\} \quad (163)$$

Then an event $E$ followed by an event $F$ can occur in

$4 \cdot 6 = 24$ ways,

$2 \cdot 6 + 4 \cdot 1 = 16$ indeterminacies of first order,

and

$2 \cdot 1 = 2$ indeterminacies of second order.

## 3.53. A Formula for the Fusion of Subjective Neutrosophic Probabilities.

For subjective neutrosophic probability, in order for us to having a better aproximation, we'd need more sources of information relating us about *the same event*. We then combine all subjective chances given by them.

Suppose a satellite is detected by radar in the sky, which can be friendly ($\underline{t}$), neutral ($\underline{i}$), or hostile ($\underline{f}$). We have two observers that give us the following subjective neutrosophic probabilities:

$$NP_1(\text{satellite}) = (t_1, i_1, f_1), \quad (164)$$

where $t_1 = ch(satellite\ is\ freindly)$,

$i_1 = ch(satellite\ is\ neutral)$,

$f_1 = ch(satellite\ is\ hostile)$,

and

$$NP_2(\text{satellite}) = (t_2, i_2, f_2). \quad (165)$$

We consider the following normalized probabilities:

$$t_1 + i_1 + f_1 = t_2 + i_2 + f_2 = 1\,, \quad (166)$$

but in case if they are non-normalized, the problem is solved in the same way. Note that $t$ stands for truth, $i$ stands for indeterminacy, and $f$ stands for falsehood.

$$(NP_1 \cap NP_2(t) = t_1 t_2 + \left(\frac{t_1^2 i_2}{t_1+i_2} + \frac{t_2^2 i_1}{t_2+i_1}\right) + \left(\frac{t_1^2 f_2}{t_1+f_2} + \frac{t_2^2 f_1}{t_2+f_1}\right). \quad (167)$$

Because:

$t_1 \cdot i_2$ is redistributed back to the truth ($t$) and indeterminacy ($i$), proportionally with respect to $t_1$, and respectively to $i_2$:

$$\frac{x_1}{t_1} = \frac{y_1}{i_2} = \frac{t_1 i_2}{t_1+i_2}, \quad (168)$$

where $x_1$ and $y_1$ are the parts from $t_1 \cdot i_2$ that are redistributed back to $t$ (chance of friendly) and respectively to $i$ (chance of neutral),

whence

$$x_1 = \frac{t_1^2 i_2}{t_1 + i_2}, \quad y_1 = \frac{t_1 i_2^2}{t_1 + i_2}. \tag{169}$$

Similarly $t_2 \cdot i_1$ is redistributed back to the truth ($t$) and indeterminacy ($i$), proportionally with respect to $t_2$, and respectively to $i_1$:

$$\frac{x_2}{t_2} = \frac{y_2}{i_1} = \frac{t_2 i_1}{t_2 + i_1}, \tag{170}$$

where $x_2$ and $y_2$ are the parts from $t_2 \cdot i_1$ that are redistributed back to $t$ (chance of friendly) and respectively to $i$ (chance of neutral),

whence

$$x_2 = \frac{t_2^2 i_1}{t_2 + i_1}, \quad y_2 = \frac{t_2 i_1^2}{t_2 + i_1}. \tag{171}$$

Again, $t_1 \cdot f_2$ is redistributed back to $t$ and $f$ (falsehood) proportionally with respect to $t_1$ and respectively $f_2$:

$$\frac{x_3}{t_1} = \frac{z_1}{f_2} = \frac{t_1 f_2}{t_1 + f_2}, \tag{172}$$

whence

$$x_3 = \frac{t_1^2 f_2}{t_1+f_2}, \quad z_1 = \frac{t_1 f_2^2}{t_1+f_2}. \tag{173}$$

And, similarly $t_2 \cdot f_1$ is redistributed back to *t* and *f* proportionally with respect to $t_2$ and respectively $f_1$:

$$\frac{x_4}{t_2} = \frac{z_2}{f_1} = \frac{t_2 f_1}{t_2+f_1}, \tag{174}$$

whence

$$x_4 = \frac{t_2^2 f_1}{t_2+f_1}, \quad z_2 = \frac{t_2 f_1^2}{t_2+f_1}. \tag{175}$$

In the same way, $i_1 \cdot f_2$ is redistributed back to *i* and *f* proportionally with respect to
$i_1$ and respectively $f_2$:

$$\frac{y_3}{i_1} = \frac{z_3}{f_2} = \frac{i_1 f_2}{i_1+f_2}, \tag{176}$$

whence

$$y_3 = \frac{i_1^2 f_2}{i_1+f_2}, \quad z_3 = \frac{i_1 f_2^2}{i_1+f_2}. \tag{177}$$

While $i_2 \cdot f_1$ is redistributed back to *i* and *f* proportionally with respect to $i_2$ and respectively $f_1$:

$$\frac{y_4}{i_2} = \frac{z_4}{f_1} = \frac{i_2 f_1}{i_2+f_1}, \tag{178}$$

whence

$$y_4 = \frac{i_2^2 f_1}{i_2+f_1}, \quad z_4 = \frac{i_2 f_1^2}{i_2+f_1}. \tag{179}$$

Then,

$$(NP_1 \cap NP_2)(i) = i_1 i_2 + \left(\frac{i_1^2 t_2}{i_1+t_2} + \frac{i_2^2 t_1}{i_2+t_1}\right) + \left(\frac{i_1^2 f_2}{i_1+f_2} + \frac{i_2^2 f_1}{i_2+f_1}\right) \tag{180}$$

and

$$(NP_1 \cap NP_2)(f) = f_1 f_2 + \left(\frac{f_1^2 t_2}{f_1+t_2} + \frac{f_2^2 t_1}{f_2+t_1}\right) + \left(\frac{f_1^2 i_2}{f_1+i_2} + \frac{f_2^2 i_1}{f_2+i_1}\right). \tag{181}$$

## 3.54. Numerical Example of Fusion of Subjective Neutrosophic Probabilities:

$$(0.6, 0.1, 0.3) \wedge_A (0.2, 0.3, 0.5) = (0.44097, 0.15000, 0.40903), \tag{182}$$

because

$$\left.\begin{matrix} t_1 = 0.6, i_1 = 0.1, f_1 = 0.3 \\ t_2 = 0.2, i_2 = 0.3, f_2 = 0.5 \end{matrix}\right\} \tag{183}$$

are replaced into the three previous neutrosophic formulas:

$$(NP_1 \cap NP_2)(t) =$$

$$= 0.6(0.2) + \left[\frac{0.6^2(0.3)}{0.6+0.3} + \frac{0.2^2(0.1)}{0.2+0.1}\right] + \left[\frac{0.6^2(0.5)}{0.6+0.5} + \frac{0.2^2(0.3)}{0.2+0.3}\right] \simeq 0.44097;$$

$$(NP_1 \cap NP_2)(i) =$$

$$= 0.1(0.3) + \left[\frac{0.1^2(0.2)}{0.1+0.2} + \frac{0.3^2(0.6)}{0.3+0.6}\right] + \left[\frac{0.1^2(0.5)}{0.1+0.5} + \frac{0.3^2(0.3)}{0.3+0.3}\right] \simeq 0.15000;$$

$$(NP_1 \cap NP_2)(f) =$$

$$= 0.3(0.5) + \left[\frac{0.3^2(0.2)}{0.3+0.2} + \frac{0.5^2(0.6)}{0.5+0.6}\right] + \left[\frac{0.3^2(0.3)}{0.3+0.3} + \frac{0.5^2(0.1)}{0.5+0.1}\right] \simeq 0.40903.$$

(184)

Therefore, there is a higher chance that the satellite is friendly, because:

$$0.44097 > 0.40903 > 0.15000. \qquad (185)$$

## 3.55. General Formula for Fusioning Classical Subjective Probabilities Provided by Two Sources.

The principle of redistributing the conflicting chances, for example $t_1 i_2$, back to $\underline{t}$ and $\underline{i}$, is the same as in PCR5 rule ( Proportional Conflict Redistribution rule #5 from *The Dezert-Smarandache Theory of Paradoxist and Plausible Reasoning (DSmT)* ) used in information fusion:

if two sources of information $S_1$, and $S_2$ give the subjective probabilities $P_1$ and $P_2$ about the same event $E$ to occur,

then combining both of them using PCR5 we get for any event E in the subjective probability space Ω,

$$(P_1 \wedge P_2)(E) = P_1(E) \cdot P_2(E) + \sum_{\substack{x \in \Omega \\ x \cap E = \emptyset}} \left[ \frac{P_1(E)^2 \cdot P_2(x)}{P_1(E) + P_2(x)} + \frac{P_2(E)^2 \cdot P_1(x)}{P_2(E) + P_2(x)} \right]. \tag{186}$$

## 3.56. Different Ways of Combining Neutrosophic Subjective Probabilities Provided by Two Sources.

Let's set the problem in a different way and use different notations. By combining subjective probabilities we don't get a single result, but many, since we do aproximations.

A military veicle is moving in a warzone. Two experts report the chances of the vehicle to be friendly (F), neutral (N), or hostile (H):

$$NP_1(\text{vehicle}) = (F_1, N_1, H_1) \quad (187)$$

and

$$NP_2(\text{vehicle}) = (F_2, N_2, H_2), \quad (188)$$

where all $F_1, N_1, H_1, F_2, N_2, H_2$ are chances (numbers in [01,]), such that $F_1 + N_1 + H_1 = F_2 + N_2 + H_2 = 1$ (normalized neutrosophic probabilities).

Suppose $NP_1 \wedge NP_2 = (F, N, H)$, with similarly $F, N, H$ in [0,1] and $F + N + H = 1$.

Let's multiply $(F_1 + N_1 + H_1)(F_2 + N_2 + H_2)$ $=1 \times 1=1$.

We get *9* terms in the left side:

$$F_1F_2 + F_1N_2 + F_1H_2 + N_1F_2 + N_1N_2 + N_1H_2 + H_1F_2 + H_1N_2 + H_1H_2 = 1. \quad (189)$$

These 9 terms are distributed to *F, N, H*.

Of course $F_1F_2$ will go to $F, N_1N_2$ wil go to *N*, and $H_1H_2$ will go to *H*.

The other 6 terms have to be distributed to *F, N, H* too.

We pay attention to the symmetry of distribution of these 6 terms to *F* and *H*.

a) *Pessimistic case*:

$F_1N_1$ and $F_2N_1$ to *N*.

Similarly $H_1N_2$ and $H_2N_1$ to *N*.

There are left $F_1H_2$ and $F_2H_1$.

a1) We can either distribute both of them to N (in a *very pessimistic case*);

a2) or we can use, for example PCR5, to redistribute them back to *F* and *H* proportionally (in a *less pessimistic way*).

b) *Optimistic case*:

$F_1N_1$ and $F_2N_1$ to *F*.

Similarly $H_1N_2$ and $H_2N_1$ to *H*.

There are left $F_1H_2$ and $F_2H_1$,

b1) We can either distribute both of them to N (in a *less optimistic case*),

b2) or we can use for example PCR5 to redistribute them back to $F$ and $H$ proportionally (in a *very optimistic way*).

No normalization in needed, since the sum $F + N + H$ will be 1.

## 3.57. Neutrosophic Logic Inference type in Fusioning Subjective Neutrosophic Probabilities.

Let the neutrosophic probability space be $\Phi = \{F, N, H\}$, where $F$ = friend, $N$ = neutral, $H$ = hostile. If we consider that all intersections of events are empty:

$$F \cap N = F \cap H = N \cap H = \emptyset, \qquad (190)$$

we can use the neutrosophic logic inference.

Suppose an aircraft is detected. We need to find out if it's a friend, or neutral, or hostile.

We have two sources that give the subjective chances:

|  | $F$ | $N$ | $H$ |
|---|---|---|---|
| $NP_1$ | $a_1$ | $a_2$ | $a_3$ |
| $NP_2$ | $b_1$ | $b_2$ | $b_3$ |

where all $a_1, a_2, a_3, b_1, b_2, b_3$ are positive and $a_1 + a_2 + a_3 = b_1 + b_2 + b_3 = 1$.

Then

$$NP_1 \wedge_p NP_2 = (a_1, a_2, a_3) \wedge_p (b_1, b_2, b_3) = (a_1 \wedge b_1, a_2 \vee b_2, a_3 \vee b_3) \quad (191)$$

in a *pessimistic way*,

or

$$NP_1 \wedge_o NP_2 = (a_1, a_2, a_3) \wedge_o (b_1, b_2, b_3) = (a_1 \wedge b_1, a_2 \wedge b_2, a_3 \vee b_3) \quad (192)$$

in an *optimistic way*.

" $\wedge$ " is a t-norm operation, and " $\vee$ " is a t-conorm operation. For example, $\wedge/\vee$ can be as in fuzzy logic respectively:

$$min/max,$$

$$xy \ / \ x + y - xy,$$

$$max\{0, x + y - 1\} \ / min\{1, x + y\}, \text{ etc.} \quad (193)$$

Then we normalize each way if needed.

The N-norm and N-conorm were defined above with the help of t-norm and t-conorm.

*Numerical Example:*

| | $F$ | $N$ | $H$ |
|---|---|---|---|
| $NP_1$ | 0.4 | 0.1 | 0.5 |
| $NP_2$ | 0.3 | 0.5 | 0.2 |

$$NP_1 \wedge_p NP_2 = (\,0.4, 0.1, 0.5) \wedge_p (0.3, 0.5, 0.2) =$$

$$= (0.4 \wedge 0.3, 0.1 \vee 0.5, 0.5 \vee 0.2) =$$

$$= (min\{0.4, 0.3\}, max\{0.1, 0.5\}, max\{0.5, 0.2\}\,) =$$

(using *min/max* operators)

$$= (0.3, 0.5, 0.5) \text{ normalized to } \left(\frac{3}{15}, \frac{5}{15}, \frac{5}{15}\right).$$

$NP_1 \wedge_0 NP_2 =$
$(min\{0.4, 0.3\}, min\{0.1, 0.5\}, max\{0.5, 0.2\}\,) =$
$(0.3, 0.1, 0.5)$ normalized to $\left(\frac{3}{9}, \frac{1}{9}, \frac{5}{9}\right)$. (194)

If we combine both pessimist and optimist results we get:

| F | N | H |
|---|---|---|
| $[\frac{3}{15}, \frac{3}{9}]$ | $[\frac{1}{9}, \frac{5}{15}]$ | $[\frac{5}{15}, \frac{5}{9}]$ |

It would be interesting of computing using the other $\wedge/\vee$ operators and compare or combine (for example making the average of) the results.

### 3.58. Neutrosophic Logic vs. Subjective Neutrosophic Probability.

In *neutrosophic logic*, the operator AND computes the conjunction of the two or more different logical propositions.

In *subjective neutrosophic probability*, there is a single neutrosophic probability space, and two or more sources of information that provide subjective chances about the events to occur. Then we use various procedures to aggregate the subjective probabilities provided by all sources in order to get the best estimation.

### 3.59. Removing Indeterminacy.

We can remove indeterminacy from the sample space, but then the second axiom of Kolmogorov is invalidated, because the neutrosophic probability of the whole sample space is strictly less than 1.

Let's suppose having a cubic die whose three faces 4, 5, 6 are erased and unreadable. We then have

$$ch(1) = ch(2) = ch(3) = \frac{1}{6} \tag{195}$$

and

$$ch(indeterm) = 3\left(\frac{1}{6}\right) = \frac{3}{6} = \frac{1}{2}. \tag{196}$$

So, if we remove the indeterminacy, our neutrosophic sample space becomes:

$$\nu\Omega = \{1, 2, 3\} \tag{197}$$

and

$$ch(\nu\Omega) = \frac{1}{2} < 1. \tag{198}$$

$\nu\Omega$ is an *incomplete classical sample space*. The first and third axioms remain valid, but the second axiom is invalided.

### 3.60. n-Valued Refined Neutrosophic Probability Space and Neutrosophic Probability.

Let's consider a handball game between two teams, Romania and Bulgaria. What is the neutrosophic probability that Romania wins?

The simplest sample space with respect to Romania is:

$$\nu\Omega = \{\text{victory}, \text{equality}, \text{defeat}\}. \quad (199)$$

Suppose that the neutrosophic probability measure *NP* (Romania wins) = (0.7, 0.1, 02). But we go further and refine the sample space and, implicitly, the probability measure.

The n-Valued Refined Neutrosophic Sample Space is:

$$\nu\Omega = \left\{\{V_1, V_2, \dots, V_p\}, \{E_1, E_2, \dots, E_h\}, \{D_1, D_2, \dots, D_s\}\right\}$$

(200)

where $p, r, s \geq 1$ and $p + r + s = n$;

also:

$V_1$ = Romania wins with 1 goal difference (i.e. 1-0, 2-1, 3-2, etc.);

..........................................................................................

$V_{p-1}$ = Romania wins with $p-1$ goals difference;

$V_p$ = Romania wins with p or more goals difference;

$E_1$ = tie game (equality), with score 0-0, 1-1;

$E_2 =$ tie game, with score 2-2;

$E_{r-1} =$ tie game, with score (r-1) to (r-1);

$E_r =$ tie game, with score *x* to *x*, where $x \geq r$.

Similarly:

$D_1 =$ Romania is defeated with 1 goal difference (i.e. 0-1, 1-2, 2-3, etc.);

..........................................................................................

$D_{s-1} =$ Romania is defeated with $s$ −1 goals difference;

$D_s =$ Romania is defeated with *s* or more goals difference.

Consequently, the *n-Valued Refined Neutrosophic Probability Measure* could be:

$$NP' \text{ (Romania wins)} =$$

$$= \left(\left(\underbrace{0.4, 0.2, 0.5, 0.05, 0, \dots, 0}_{p}\right), \left(\underbrace{0.03, 0.05, 0.02, 0 \dots, 0}_{r}\right), \left(\underbrace{0.1, 0.08, 0.02, 0, \dots, 0}_{s}\right)\right).$$

(201)

In general, let's consider a neutrosophic probability space, and a neutrosophic event $A$.

$$NP(A) = \ (ch(A), ch(indeterm_A), ch(\bar{A})) =$$

by notation to (*T, I, F*),

where for simplicity $T$ = truth, $I$ = indeterminacy, and $F$ = falsity as in neutrosophic logic.

Then the *n-Valued Refined Neutrosophic Probability* is:

$$NP_h(A) = \left((T_1, T_2, \dots, T_p), (I_1, I_2, \dots, I_r), (F_1, F_2, \dots, F_s)\right) \quad (202)$$

with $p, r, s \geq 1$ and $p + r + s = n$;

and $T_j$ = the chance that event $A$ occurs and the occurence has the property $P_j$;

$I_k$ = the indeterminacy related to the occurence of event $A$, such that the indeterminacy has the property $Q_r$;

$F_l$ = the chance that event A does not occur and the non-occurence has the property $R_l$;

where $1 \leq j \leq p, 1 \leq k \leq r,$ and $1 \leq l \leq s$.

*Remarks.*

a) Such n-Valued refinement is not possible for all applications.

a1) For example, if we consider a cubic die with four erased faces 1, 2, 3, and 4, tossed on a regular surface, then $NP(5) = \left(\frac{1}{6}, \frac{4}{6}, \frac{1}{6}\right)$; but *we are not able to refine* the occurrence of $\{5\}$, neither the non-occurrence of $\{5\}$, nor the indeterminacy that might be related to this event.

a2) But if we consider a regular die tossed on an irregular surface with several small cracks and other deep cracks, *we may refine the indeterminacy* for each event, since the die may get stucked in a small crack with a vertex of faces, we can read (for example 4 & 5 & 6), or the die can fall in a deep crack that we are not able to see it at all. Yet, we are not able to refine the occurrence or non-occurrence of an event in this case.

b) The refinements can be done in multiple ways, depending on the properties $P_j, Q_k, R_l$ we choose, for all *j, k, l*.

### 3.61. Neutrosophic Markov Chain.

It is a straight neutrosophic generalization of the classical Markov chain, i.e. some indeterminacy is taken into consideration in the classical probability space.

The neutrosophic Markov chain is a sequence of neutrosophic random variables $X_1, X_2, \ldots$, with the property that the next neutrosophic state depends on the current neutrosophic state only:

$$NP(X_n = x | X_1 = x_1, X_2 = x_2, \ldots, X_{n-1} = x_{n-1}) = NP(X_n = x | X_{n-1} = x_{n-1}). \quad (203)$$

It is a *neutrosophic mathematical system* that is characterized as memoryless.

A *neutrosophic transition* is a change of the state of a system with indeterminacy.

A *neutrosophic Markov chain of order m*, where $1 \leq m < \infty$, or neutro-sophic Markov chain with memory *m*, is:

$$NP(X_n = x | X_{n-1} = x_{n-1}, X_{n-2} = x_{n-2}, \ldots, X_1 = x_1) =$$

$$= NP(X_n = x | X_{n-1} = x_{n-1}, X_{n-2} = x_{n-2}, \ldots, X_{n-m} = x_{n-m}). \quad (204)$$

We defined above the neutrosophic Markov chain for *discrete time*. For a *continuous time*, we use a continuous index:

$$NP(X_n = x | X_{n-1} = y) = NP(X_{n-1} = x | X_{n-2} = y), \text{ for all } n. \quad (205)$$

To illustrate an example about a discrete neutrosophic Markov chain, we use a *neutrosophic probability graph* (this should be distinguished from the *neutrosophic graph* and *neutrosophic fuzzy graph*, both introduced by W.B. Vasantha Kandasamy & F. Smarandache in our algebraic structure books since year 2003).

Let's consider the world economy, and its states: economic prosperity ($P$), economic recession ($R$), and economic depression ($D$).

Suppose we have the following neutrosophic probability graph, during a year:

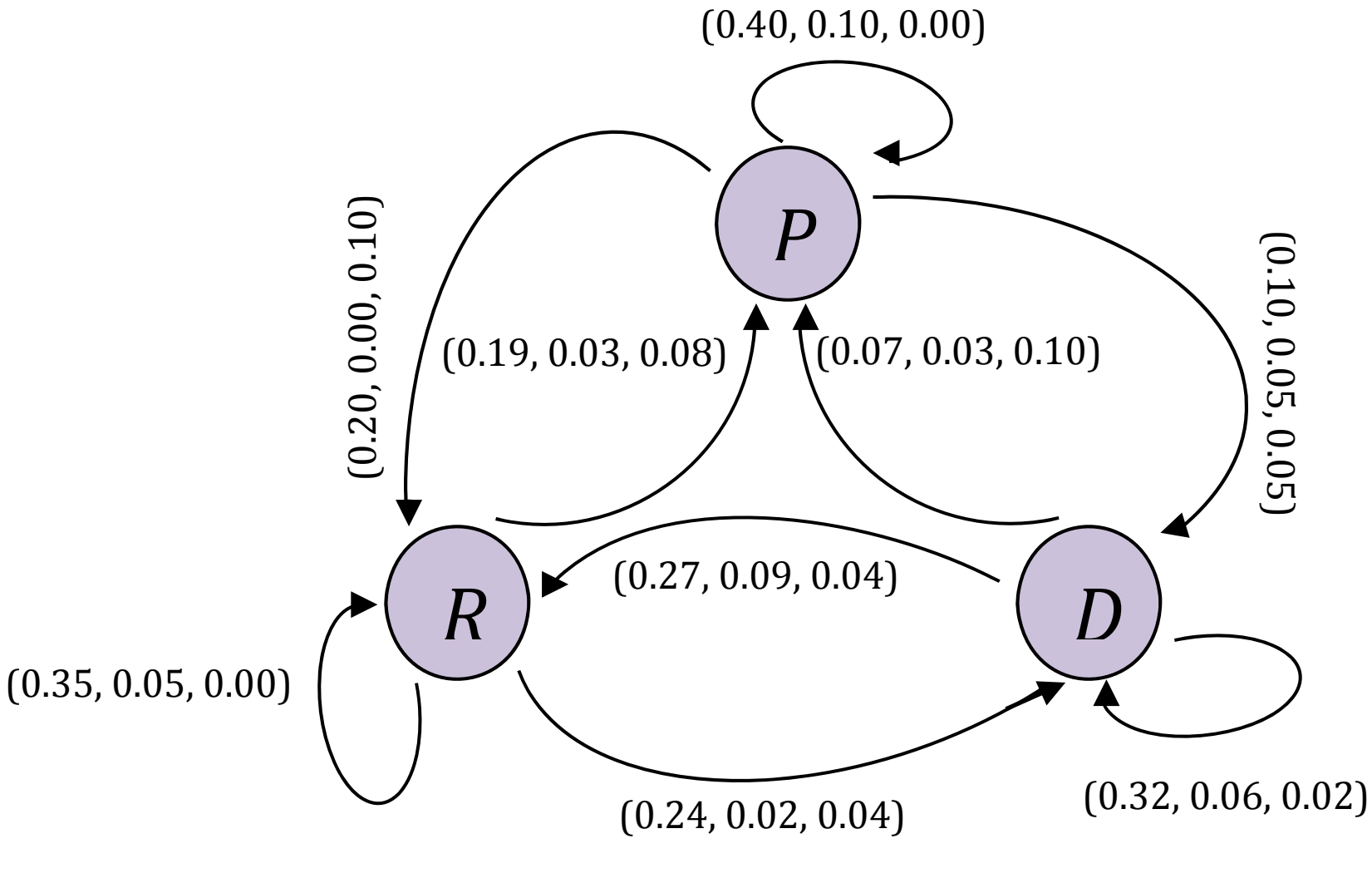

*Fig. 10*

Upon the figure, an economic prosperity year is followed by another economic prosperity year 40% of the time while 10% of the time it is unknown, an economic recession year 20% of the time, while 10% of the time it is not followed by an economic recession, and an economic depression year 10% of the time and 5% of the time it is unknown while 5% of the time it is not followed by an economic recession year.

The neutrosophic transition matrix of this graph is:

$$\begin{array}{cc} & \begin{matrix} P & & R & & D \end{matrix} \\ NP = \begin{matrix} P \\ R \\ D \end{matrix} & \begin{bmatrix} (0.40, 0.10, 0.00) & (0.20, 0.00, 0.10) & (0.10, 0.05, 0.05) \\ (0.19, 0.03, 0.08) & (0.35, 0.05, 0.00) & (0.24, 0.03, 0.04) \\ (0.07, 0.03, 0.10 & (0.27, 0.09, 0.04) & (0.32, 0.06, 0.02) \end{bmatrix} \end{array}$$

The state space is $\{P, R, D\}$.

The stochastic row vectors are:

$$\begin{aligned} P &= [1 \quad 0 \quad 0], \\ R &= [0 \quad 1 \quad 0], \\ D &= [0 \quad 0 \quad 1]. \end{aligned}$$

Let $X$ be any of these stochastic row vectors,

with the neutrosophic relation

$$X^{(n+1)} = X^{(n)} NP \qquad (206)$$

for any time $n$.

Whence

$$X^{(n+2)} = X^{(n+1)}NP$$

$$= \left[X^{(n)}NP\right]NP$$

$$= X^{(n)}(NP)^2 \qquad (207)$$

and more general:

$$X^{(n+m)} = X^{(n)}(NP)^m. \qquad (208)$$

At the end we normalize the rows of matrix $(NP)^m$.

We define the multiplication of neutrosophic probabilities as

$$(a_1, b_1, c_1) \cdot (a_2, b_2, c_2) = (a_1 a_2, max\{b_1, b_2\}, max\{c_1, c_2\}), \qquad (209)$$

and the addition of neutrosophic probabilities as:

$$(a_1, b_1, c_1) + (a_2, b_2, c_2) = (a_1 + a_2, min\{b_1, b_2\}, min\{c_1, c_2\}). \qquad (210)$$

In case when the neutrosophic probability is reduced to classical probability (i.e. $b_1 = b_2 = c_1 = c_2 = 0$), we get the same result for neutrosophical prbability matrix $NP^n$ as for the classical probability matrix $P^n$.

Other multiplication and addition operators of neutrosophic probabilities can be defined as well. For example:

$$(a_1, b_1, c_1) \cdot (a_2, b_2, c_2) = (a_1 a_2, min\{b_1, b_2\}, max\{c_1, c_2\}), \quad (211)$$

or

$$(a_1, b_1, c_1) \cdot (a_2, b_2, c_2) = \left(a_1 a_2, \frac{b_1 + b_2}{2}, max\{c_1, c_2\}\right), \text{etc.} \quad (212)$$

and similarly for the addition of neutrosophic probabilities changing the middle component to $max\{b_1, b_2\}$, or average $\frac{b_1+b_2}{2}$, etc.

Let's note $(NP)^2 = \{c_{ij}\}_{i.j}$; (213)

$$c_{11} = [(0.40, 0.10, 0.00)(0.20, 0.00, 0.10)(0.10, 0.05, 0.05)] \cdot \begin{bmatrix} (0.40, 0.10, 0.00) \\ (0.19, 0.03, 0.08) \\ (0.07, 0.03, 0.10) \end{bmatrix} = \underset{P \to P}{(0.40, 0.10, 0.00)} \cdot$$

$$\underset{P \to P}{(0.40, 0.10, 0.00)} + \underset{P \to R}{(0.20, 0.00, 0.10)} \underset{R \to P}{(0.19, 0.03, 0.08)} + \underset{P \to D}{(0.10, 0.05, 0.05)} \underset{D \to P}{(0.07, 0.03, 0.10)} =$$

$(0.16, 0.10, 0.00) + (0.038, 0.03, 0.10) +$
$(0.007, 0.05, 0.10) = (0.205, 0.050, 0.000).$ (214)

$c_{11}$ means the neutrosophic probability of having an economic prosperity year (*P*), after 2 years, starting from prosperity:

$[(P \to P) \wedge (P \to P)]$ or $[(P \to R) \wedge (R \to$
$P)]$ or $[(P \to D) \wedge (D \to P)]$, (215)

where, for example, $P \to R$ means "from prosperity to recession", and so on.

$$c_{12} = (0.080, 0.10, 0.10) + (0.070, 0.05, 0.10) + (0.027, 0.09, 0.05) = (0.277, 0.050, 0.050)$$

$$c_{12} = (0.004, 0.10, 0.05) + (0.048, 0.02, 0.10) + (0.032, 0.06, 0.05) = (0.084, 0.020, 0.050)$$

$$c_{21} = (0.076, 0.10, 0.08) + (0.0665, 0.05, 0.08) + (0.0168, 0.03, 0.10) = (0.1593, 0.003, 0.080)$$

$$c_{22} = (0.038, 0.03, 0.10) + (0.1225, 0.05, 0.00) + (0.0648, 0.09, 0.004) = (0.253, 0.003, 0.000)$$

$$c_{23} = (0.019, 0.05, 0.08) + (0.0840, 0.05, 0.04) \\ + (0.0768, 0.06, 0.04) \\ = (0.1798, 0.005, 0.004)$$

$$c_{31} = (0.028, 0.100, 0.100) \\ + (0.0513, 0.09, 0.08) \\ + (0.024, 0.060, 0.100) \\ = (0.1017, 0.06, 0.08)$$

$$c_{32} = (0.014, 0.03, 0.10) + (0.0945, 0.09, 0.04) \\ + (0.0864, 0.09, 0.04) \\ = (0.1949, 0.03, 0.04)$$

$$c_{33} = (0.007, 0.05, 0.10) + (0.0648, 0.09, 0.04) \\ + (0.1024, 0.06, 0.02) \\ = (0.1742, 0.05, 0.02).$$

Thus

$$(NP)^2 \\ = \begin{bmatrix} (0.205, 0.05, 0.0) & (0.277, 0.05, 0.05) & (0.084, 0.02, 0.05) \\ (0.1593, 0.03, 0.08) & (0.2253, 0.03, 0.0) & (0.1798, 0.05, 0.04) \\ (0.1017, 0.06, 0.08) & (0.1949, 0.03, 0.04) & (0.1742, 0.05, 0.02) \end{bmatrix}. \tag{216}$$

We normalize the rows by dividing each of the nine components by their sum. We get with three decimal approximation:

$$(NP)^2_{norm} = \begin{bmatrix} (0.261, 0.064, 0.000) & (0.352, 0.064, 0.064) & (0.106, 0.025, 0.064) \\ (0.201, 0.038, 0.101) & (0.284, 0.038, 0.000) & (0.226, 0.062, 0.050) \\ (0.135, 0.080, 0.107) & (0.260, 0.040, 0.053) & (0.232, 0.066, 0.027) \end{bmatrix}.$$

(217)

According to this *neutrosophic transition matrix*, after two years the largest chance of the economy to be is in the state of recession.

### 3.62. Applications of Neutrosophics.

Once could use the neutrosophics in statistical physics, financial markets, risk management, mathematical biology, quantum theory, and in almost any humanistic or scientific field where indeterminacy, unknown, and in general where <neutA> (neutrality with respect to an item <A>) occur.

# Chapter 4.
# Neutrosophic Subjects for Future Research

## Neutrosophic Subjects

1. Neutrosophic topologies, including neutrosophic metric spaces and smooth topological spaces.
2. Neutrosophic numbers ($a+bI$, where $I$ = indeterminacy and $I^2 = I$, $mI+nI = (m+n)I$, $0I = 0$, and $a$, $b$ are real or complex numbers), and arithmetical operations, including ranking procedures for neutrosophic numbers.
3. Neutrosophic rough sets.
4. Neutrosophic relational structures, including neutrosophic relational equations, neutrosophic similarity relations, and neutrosophic ordering.
5. Neutrosophic geometry (Smarandache geometries).
6. Neutrosophic probability.
7. Neutrosophic logical operations, including n-norms, n-conorms, neutrosophic implicators, neutrosophic quantifiers.
8. Measures of neutrosophication.
9. Deneutrosophication techniques.
10. Neutrosophic multivalued mappings.
11. Develop the neutrosophic measure (defined in this book).
12. Develop the neutrosophic integral (defined in this book).
13. Neutrosophic differential calculus.
14. Neutrosophic mathematical morphology.
15. Neutrosophic algebraic structures.
16. Neutrosophic models.
17. Neutrosophic cognitive maps.

18. Neutrosophic relational maps.
19. Neutrosophic matrix, bimatrix, ..., $n$-matrix.
20. Neutrosophic graph, which is a graph that has at least one indeterminate edge or one indeterminate node.
21. Neutrosophic tree, which is a tree that has at least one indeterminate edge or one indeterminate node.
22. Neutrosophic fusion rules for information fusion.
23. Applications: neutrosophic relational databases, neutrosophic image (thresholding, denoising, segmentation) processing, neutrosophic linguistic variables, neutrosophic decision making and preference structures, neutrosophic expert systems, neutrosophic reliability theory, neutrosophic soft computing techniques in e-commerce and e-learning, image segmentation, etc.

# ADDENDA

## Books on Neutrosophics

## More Articles on Neutrosophics

## Seminars on Neutrosophics

1. *An Introduction to Information Fusion Level 1 and to Neutrosophic Logic/Set with Applications*, by F. Smarandache, ENSIETA (National Superior School of Engineers and the Study of Armament), Brest, France, 2 July 2010.
2. *An Introduction to Fusion Level 1 and to Neutrosophic Logic/Set with Applications*, by F. Smarandache at Air Force Research Laboratory, in Rome, NY, USA, July 29, 2009.
3. *An Introduction to Neutrosophic Logic in Arabic Philosophy*, by F. Smarandache & Salah Osman, Minufiya University, Shebin Elkom, Egypt, 17 December 2007.

## International Conferences on Neutrosophics

1. International Conference on Applications of Plausible, Paradoxical, and Neutrosophic Reasoning for Information Fusion, Cairns, Queensland, Australia, 8-11 July 2003.
2. First International Conference on Neutrosophy, Neutrosophic Logic, Set, Probability and Statistics, University of New Mexico, Gallup Campus, 1-3 December 2001.

## Ph. D. Dissertations on Neutrosophics

1. Eng. Ionel Alexandru Gal, *Contributions to the Development of Hybrid Force-Position Control Strategies for Mobile Robots Control*, advisers Dr. Luige Vlădăreanu & Dr. Florentin Smarandache, Institute of Solid Mechanics, Romanian Academy, Bucharest, October 14, 2013.
2. Smita Rajpal, *Intelligent Searching Techniques to Answer Queries in RDBMS, Ph D Dissertation in progress*, under the supervision of Prof. M. N. Doja, Department of Computer Engineering Faculty of Engineering, Jamia Millia Islamia, New Delhi, India, 2011.

3. Ming Zhang, *Novel Approaches to Image Segmentation Based on Neutrosophic Logic*, *Ph D Dissertation*, Utah State University, Logan, Utah, USA, All Graduate Theses and Dissertations, Paper 795, http://digitalcommons.usu.edu/etd/795, 12-1-201, 2010.